\documentclass{article} %
\usepackage{iclr2026_conference,times}

\usepackage{amsmath,amsfonts,bm}

\def\eqref#1{equation~\ref{#1}}

\def\1{\bm{1}}

\DeclareMathAlphabet{\mathsfit}{\encodingdefault}{\sfdefault}{m}{sl}
\SetMathAlphabet{\mathsfit}{bold}{\encodingdefault}{\sfdefault}{bx}{n}

\usepackage{booktabs, tabularx, colortbl, array}
\usepackage{algpseudocode}
\usepackage{algorithm}
\usepackage[colorlinks=true]{hyperref}
\usepackage[capitalize,noabbrev]{cleveref}
\usepackage{url}
\usepackage{graphicx}
\usepackage{tikz}            
\usetikzlibrary{positioning}
\usepackage{longtable}
\usepackage{enumitem}
\usepackage{amsmath,amssymb}
\usepackage{capt-of}
\usepackage{todonotes}
\usepackage[most]{tcolorbox}
\usepackage{etoolbox}

\usepackage{color}
\definecolor{cgreen}{rgb}{0.2,0.6,0.2}
\definecolor{darkred}{rgb}{0.4,0.0,0.0}
\definecolor{darkgreen}{rgb}{0.0,0.4,0.0}
\definecolor{darkblue}{rgb}{0.0,0.0,0.4}

\hypersetup{
  linkcolor = black,     %
  citecolor = darkblue,  %
  urlcolor  = darkblue   %
}

\iclrfinalcopy

\title{Noise-Response Calibration: A Causal Intervention Protocol for LLM-Judges}

\author{Maxim Khomiakov$^{1,2,3}$ \& Jes Frellsen$^{1,3}$ \\
$^{1}$Technical University of Denmark, Kgs. Lyngby, Denmark \\
$^{2}$Normal Computing Corporation, New York, NY, USA \\
$^{3}$Pioneer Centre for Artificial Intelligence, Copenhagen, Denmark\\
\texttt{maxim@normalcomputing.com}, \texttt{\{maxk,jefr\}@dtu.dk}
}

\begin{document}
\maketitle

\begin{abstract}
Large language models (LLMs) are increasingly used as automated judges and synthetic labelers, especially in low-label settings. Yet these systems are stochastic and often overconfident, which makes deployment decisions difficult when external ground truth is limited. We propose a practical calibration protocol based on controlled input interventions: if noise severity increases, task performance should exhibit a statistically significant deterioration trend. We operationalize this with a slope-based hypothesis test over repeated trials, using signal-to-noise-ratio (SNR) perturbations for tabular data and lexical perturbations for text data. Across UCI tabular benchmarks and four text classification datasets, we find clear modality-dependent behavior. Our results reveal a modality gap: while text-based judges degrade predictably, the majority of tabular datasets show a lack of statistically significant performance deterioration even under significant signal-to-noise reduction. Interestingly we find that model performance is lower on datasets that are insensitive to noise interventions. We present a reproducible methodology and reporting protocol for robust LLM-judge calibration under distribution shift.
\end{abstract}

\section{Introduction}
High-quality labels remain costly and difficult to produce: annotation policy design,
expert recruitment, and quality control are all substantial bottlenecks
\citep{gebru2021datasheetsdatasets,snow-etal-2008-cheap}. In parallel, LLM-based
pipelines now support two high-value workflows: (1) synthetic labeling, and (2)
LLM-as-a-Judge decision systems for automated triage and ranking
\citep{liu2023gevalnlgevaluationusing,zheng2023judgingllmasajudgemtbenchchatbot}.

The practical failure mode is not only low average quality, but unreliable behavior
under shift: small input changes can create non-trivial output volatility, especially
when large trusted holdout labels are unavailable
\citep{lin2022truthfulqameasuringmodelsmimic,huang2024surveyhallucinationlargelanguage}.

\paragraph{Core question.}
Can we reject an LLM judge for a target task \emph{before} deployment by stress-testing
its response to controlled interventions? We frame this as a causal question in the interventionist sense of \citet{pearl2009causality}: apply a controlled perturbation to the input and test whether performance degrades as predicted.

We formalize a monotone-deterioration hypothesis over a calibrated noise schedule. Let
$\mathcal{P}$ denote a task-specific performance score (e.g., accuracy for classification, $R^2$ for regression) measured on clean inputs, and $\mathcal{P}_k$ the same score measured under
perturbation magnitude $k \in K$, where $K$ is ordered from weak to strong noise.
Define $\Delta_k=\mathcal{P}_k-\mathcal{P}$. Our conceptual intervention hypothesis is\looseness=-1
\begin{align}
    H_0 &: \exists\, k_1 < k_2 \text{ such that }
    \mathbb{E}[\mathcal{P}_{k_2}] > \mathbb{E}[\mathcal{P}_{k_1}],
    \label{eq:h0-monotonicity}\\
    H_1 &: \mathbb{E}[\mathcal{P}_{k_1}] \ge \mathbb{E}[\mathcal{P}_{k_2}]
    \quad \forall k_1 < k_2,
    \label{eq:h1-monotonicity}
\end{align}
with at least one strict inequality under $H_1$.

Here, $H_0$ (the null hypothesis) states that the expected performance is \emph{not}
monotone under increasing perturbation i.e., there exists at least one pair of
noise magnitudes where performance improves despite stronger noise. $H_1$ (the
alternative) encodes the expected deterioration property: as noise increases, the
expected performance does not increase, and it decreases for at least one step.
The expectation $\mathbb{E}[\mathcal{P}_k]$ is taken over the randomness in the
procedure at severity level $k$, including (i) the sampled perturbations/noise
realizations and (ii) any stochasticity in the LLM inference; In practice, we operationalize this ordered alternative using a one-sided linear trend
test (negative slope) as a pragmatic surrogate; thus sensitive indicates significant
deterioration with increasing noise, which is consistent with but not equivalent to strict
monotonicity.

Intuitively, if increasing perturbation does \emph{not} induce expected deterioration,
the model is either exploiting artifacts or is insufficiently sensitive to task signal,
both of which reduce trust in judge-style deployment. Conversely, when the expected
deterioration holds with statistical significance, we obtain a useful calibration
signal about where the model can be trusted.

This paper presents a reproducible methodology around this idea: a pipeline for
perturbation design, repeated inference, and hypothesis testing across tabular and
text settings, presenting findings which indicate that employing such a methodology prior to deployment could provide an indicator of trustworthy LLM results.

\paragraph{Overview of Contributions} 
\begin{enumerate}[leftmargin=*,itemsep=2pt]
  \item We propose a practical \emph{noise-response calibration protocol} for deciding
  whether an LLM judge is reliable for a specific task, based on intervention schedules
  and slope-based hypothesis testing.
  \item We report large-scale empirical findings spanning UCI tabular tasks and four
  widely used text classification datasets, performing experiments applying both correlated and uncorrelated
  Gaussian noise for tabular data and lexical noise for text.
  \item We demonstrate that for the majority of included UCI tabular tasks, performance is largely \emph{insensitive} to the tested noise schedules (i.e., it does not exhibit statistically significant deterioration as noise increases). We discuss how this insensitivity can serve as a practical warning signal for LLM-judge deployment, complementing average-performance reporting.
\end{enumerate}

\section{Related works}
\paragraph{Hallucination, alignment, and judge reliability.}
Recent work has studied hallucination mechanisms and mitigation strategies in depth,
including retrieval augmentation, verification, and calibration-oriented prompting
\citep{huang2024surveyhallucinationlargelanguage}. In parallel, alignment methods such
as RLHF and constitutional training have improved instruction-following behavior, but
do not eliminate robustness failures under distribution shift
\citep{ouyang2022training,bai2022constitutional}.

For LLM-as-a-Judge pipelines, reporting and evaluation protocol quality is itself a
major source of variance. A central recommendation in recent methodology work is that
judge results should be reported with stronger controls for prompt sensitivity,
variance, and statistical uncertainty \citep{lee2025llmjudge}. 

\paragraph{Perturbation sensitivity in reasoning and NLP.}
Our work is closely related to perturbation-based robustness analysis. In mathematical
reasoning, \citet{mirzadeh2024gsmsymbolic} show that seemingly benign symbolic
variations can induce substantial performance changes. More broadly, LLM reasoning has
been shown to degrade under distractor-like or superficially irrelevant perturbations
\citep{shi2023distracted}.

For text corruption, lexical perturbation studies demonstrate that character-level
noise can strongly hurt model behavior even when semantics are mostly preserved
\citep{belinkov2018synthetic,pruthi2019combating}. Our text intervention family is
inspired by this line of work but is used for \emph{trust calibration} rather than
adversarial training.

\paragraph{Causal framing and benchmark context.}
We adopt an intervention-based causal framing: the perturbation operator is treated as
an intervention on inputs, and the downstream performance response is the measured
effect \citep{pearl2009causality,peters2017elements}. This perspective makes explicit
what is being manipulated and what constitutes expected behavior under increasing
uncertainty.

For tabular benchmarks, we use datasets from the UCI repository
\citep{Asuncion2007}, one of the largest tabular datasets designed for ML prediction tasks, and include canonical datasets such as Iris
\citep{fisher1936iris} and Wine Quality \citep{cortez2009wine}. For text experiments
we use IMDB \citep{maas2011imdb}, Yelp Review Full \citep{zhang2015character}, SST-2
\citep{socher2013recursive}, and Financial PhraseBank
\citep{malo2014gooddebt}, datasets that all revolve around sentiment classification tasks.

\section{Method}\label{sec:method}

\begin{algorithm}[t]
\caption{Noise-response calibration protocol}
\label{alg:noise-calibration}
\begin{algorithmic}[1]
\Require Dataset $D$, noise schedule $N=\{n_k\}_{k=1}^K$, repetitions $R$, significance level $\alpha$
\State Initialize LLM Judge for the prediction task and compute baseline performance $\mathcal{P}$
\For{each $n_k \in N$}
  \For{each repetition $r \in \{1,\dots,R\}$}
    \State Generate noisy inputs, query the LLM, record performance $\mathcal{P}_{k,r}$
  \EndFor
\EndFor
\State Fit linear model $\mathcal{P}_{k,r} = \beta_0 + \beta_1 n_k + \epsilon$
\State Perform one-sided slope test $H_0: \beta_1 \ge 0$ vs.\ $H_1: \beta_1 < 0$
\State Accept task-level trust calibration only if deterioration is significant and directionally consistent
\end{algorithmic}
\end{algorithm}

Our procedure is conceptually close to mutation testing in software engineering where one applies
controlled perturbations, observes failure behavior, and decides whether a system should
be trusted for deployment-critical use \citep{demillo1978hints,jia2011analysis}. Designing a sufficiently general noise family is non-trivial and domain-dependent. The
intervention family can miss realistic corruption patterns, so acceptance should be
interpreted as conditional on the tested noise family, not as a universal robustness claim.
We treat this explicitly as a limitation and future direction (Section~\ref{sec:future}).

For each dataset, we build a dynamic system prompt containing task explanation, dataset metadata, output constraints, and provide few shot examples as part of the input.
Predictions are validated through a schema-constrained output parser, where one requires the answer to match the format of the label. We present our noise-response calibration protocol in Algorithm~\ref{alg:noise-calibration}.

\subsection{Tabular noise via SNR schedules}
For tabular inputs, let $X_{\mathrm{clean}}\in\mathbb{R}^{N\times d}$ denote the full
clean training matrix restricted to numeric covariates, and let
$x\in\mathbb{R}^d$ denote one evaluation row. We estimate the reference covariance
once on the full clean training data,
$\Sigma_{\text{signal}} = \mathrm{Cov}(X_{\mathrm{clean}})$, and write
$\sigma_j^2 = \Sigma_{\text{signal}, jj}$ for the empirical
variance of feature $j$. Let $D = \mathrm{diag}(\sigma_1^2,\dots,\sigma_d^2)$ and
$R = \mathrm{Corr}(X_{\mathrm{clean}})$, so that
$\Sigma_{\text{signal}} = D R D$. For a target $\mathrm{SNR}_{\mathrm{dB}}$, define
$\alpha = 10^{-\mathrm{SNR}_{\mathrm{dB}}/10}$. This scales the noise power
relative to the fixed clean-data signal statistics, and we evaluate a fixed ordered
schedule from mild to strong perturbation.

Our interventions operate only on numeric covariates; datasets with low numeric-feature
coverage receive partial or no perturbation, and non-numeric covariates are preserved.
We add Gaussian noise $x_{\text{noisy}} = x + \epsilon$ under two covariance structures:
\begin{enumerate}[leftmargin=*,itemsep=1pt,topsep=2pt]
    \item \textbf{Uncorrelated}: $\epsilon \sim \mathcal{N}(0,\,\alpha D^2)$, equivalently
    $\epsilon \sim \mathcal{N}(0,\,\alpha\,\mathrm{diag}(\sigma_1^2,\dots,\sigma_d^2))$.
    \item \textbf{Correlated}: $\epsilon \sim \mathcal{N}(0,\,\alpha \Sigma_{\text{signal}})$,
    equivalently $\epsilon \sim \mathcal{N}(0,\,\alpha D R D)$.
\end{enumerate}
Both perturbations are zero-mean and share the same per-feature marginal variances
$\alpha \sigma_j^2$; they differ only in whether cross-feature covariance is
suppressed (uncorrelated) or preserved according to the empirical correlation
structure of the full clean training data (correlated). This lets us test whether
sensitivity depends only on marginal noise scale or also on the directional structure
of the perturbation.

\subsection{Text datasets}
We evaluate IMDB, Yelp Review Full, SST-2, and Financial PhraseBank
\citep{maas2011imdb,zhang2015character,socher2013recursive,wang2019glue,malo2014gooddebt}.
All are classification tasks, but they differ in domain, label granularity, and text
length.

\paragraph{Noise intervention design.}
For text, we focus on lexical corruption inspired by prior robust and adversarial NLP work
\citep{belinkov2018synthetic,pruthi2019combating}. While prior work primarily uses such perturbations to
\emph{train} or \emph{harden} models against corrupted inputs, we leverage the same mechanisms to
evaluate whether an LLM judge exhibits the expected degradation as corruption increases.

\paragraph{Lexical noise.}
Let $x=(w_1,\dots,w_T)$ be a tokenized input and $\alpha\in[0,1]$ a severity control.
We define
\begin{align}
    x_{\text{noisy}} \sim \mathcal{C}_\alpha(x),
\end{align}
where $\mathcal{C}_\alpha$ applies token-level corruptions with probability
\begin{align}
    p(\alpha)=\alpha p_{\max}.
\end{align}
For each token $w_i$, we sample whether to corrupt, then sample an operation:
word dropout, adjacent character swap, keyboard-adjacent typo, or random
insertion/deletion relative to the corruption severity level $\alpha$.

\subsection{Deterioration analysis}
For each setting (dataset $\times$ noise type/intensity), we fit an ordinary least-squares regression
\begin{align}
    \mathcal{P}_{k,r} = \beta_0 + \beta_1 n_k + \epsilon_{k,r},
\end{align}
where $k\in\{1,\dots,K\}$ indexes the noise severity level, $r\in\{1,\dots,R\}$ indexes the independent repetition at that level, $\mathcal{P}_{k,r}$ is the observed performance score for severity $k$ and repetition $r$, $\beta_0$ is the intercept (expected performance at zero added noise), $\beta_1$ is the slope measuring the rate of performance change per unit of noise intensity, $n_k$ is linear in true noise intensity (for tabular data,
$n_k=\mathrm{SNR}_{\max}-\mathrm{SNR}_k$), and $\epsilon_{k,r}\overset{\text{iid}}{\sim}\mathcal{N}(0,\sigma^2_\epsilon)$ is the residual error term. We then test
\begin{align}
    H_0 &: \beta_1 \geq 0, \\
    H_1 &: \beta_1 < 0, \label{eq:snr_x_onesided_test}
\end{align}
at significance level $\alpha=0.05$, using the standard OLS $t$-statistic for the slope coefficient \citep{seber2012linear}:
\begin{align}
    t = \frac{\hat{\beta}_1}{\mathrm{SE}(\hat{\beta}_1)},
    \qquad
    t \sim \text{Student-}t_{N-2} \text{ under } H_0,
\end{align}
where $\hat{\beta}_1$ is the OLS slope estimate, $\mathrm{SE}(\hat{\beta}_1)$ is its standard error, and $N = K \times R$ is the total number of observations. We reject $H_0$ (and label the task \textit{sensitive}) when $t < t_{\alpha, N-2}$, where $t_{\alpha, N-2}$ is the critical value corresponding to significance level $\alpha$ for a Student's $t$-distribution
with $N-2$ degrees of freedom. That is when the observed slope is significantly negative at the chosen level. This operational test approximates the ordered alternative in \cref{eq:h0-monotonicity}--\cref{eq:h1-monotonicity} via a one-sided negative-trend surrogate. Thus, for all the datasets we compute a one-directional test of negative slope, to inform us as to whether the dataset in question proved \textit{sensitive} or \textit{insensitive} to the noise intervention.

\section{Experimental Setup}
We run all main experiments with OpenAI \textsc{GPT-5-mini} as the base judge model. For each dataset and noise level, we run 5 repetitions and provide 20 few-shot examples before recording the predicted value. Our goal is to give the model the best possible basis for solving the task at hand before deliberately handicapping its odds. For the prompt serialization we rely on a fairly simple approach, providing a dataset and task description, information about the label and output format constraints. Further details on the prompt leveraged is described in Appendix \cref{box:prompt-setup}.

\subsection{Datasets and tasks}
Our benchmark covers three domains:
\begin{enumerate}[leftmargin=*]
  \item UCI classification: 222 datasets.
  \item UCI regression: 143 datasets.
  \item Text classification: 4 datasets (IMDB, Yelp Review Full, SST-2,  Financial
  PhraseBank).
\end{enumerate}

All our datasets are split into train/valid/test fractions
$0.70/0.15/0.15$, with stratified training split where applicable. Few shot examples are taken from the training set.

\subsection{UCI eligibility and filtering}
\label{sec:uci-eligibility}
As previously mentioned, we rely on additive Gaussian noise and therefore apply it only to numeric features. Not all datasets, however, have sufficiently many numerical features available. We therefore use explicit exclusion criteria: (i) a minimum proportion of numeric features, and (ii) removal of datapoints with missing values. If these filters leave too few observations, we skip the dataset. We summarize the resulting dataset eligibility flow in Table~\ref{tab:uci-exclusion-flow}.

This filtering reflects a common practical limitation in tabular-LLM evaluation pipelines, where (i) numeric-only perturbations do not apply to categorical-heavy datasets and (ii) prompt-length constraints limit how much structured information can be provided to the model \citep{fang2024llmstabular,pmlr-v206-hegselmann23a,sui2024tablemeetsllm}.

\begin{table}[t]
\caption{Stepwise UCI exclusion flow. First filter keeps datasets with $>50\%$ numerical
covariates; second filter applies missing-value incomputability rules; the final row reports the
subset used for per-dataset slope analyses in Appendix
Table~\ref{tab:uci-all-slopes}.}
\label{tab:uci-exclusion-flow}
\centering
\small
\begin{tabular}{p{0.62\linewidth}cc}
\toprule
Step / criterion & Classification & Regression \\
\midrule
Benchmark inventory declared in this paper & 222 & 143 \\
After $>50\%$ numeric feature filter & $190$ ($-32$) & $127$ ($-16$) \\
After missing-value drop-off & $185$ ($-5$) & $125$ ($-2$) \\
Complete datasets included in this study & 138 & 21 \\
\bottomrule
\end{tabular}
\end{table}

The final count of 138 classification and 21 regression datasets reflects those for which we obtained complete experimental results across all noise levels and repetitions. Extending coverage to the remaining eligible datasets is discussed in Section~\ref{sec:future}.

\paragraph{Performance metrics and configuration}
For the performance metrics we compute confusion metrics such as F1, precision and recall but rely on accuracy as the primary measure for the classification studies. Similarly, we compute Mean Squared Error, Mean Absolute Error for each regression task while relying on $R^2$ as the primary measure for the regression studies. Unless explicitly stated otherwise, we use $n_{\text{context}}=20$ few shot examples per
prompt. To control prompt length in tabular settings, we cap the number of features
shown per example (10 features maximum). For each level of noise we repeat the experiment with five
independent trials. Finally, evaluation instances are processed in batches with 500 rows for tabular datasets and 50 examples for text.

Upon concluding the experiments for each dataset, we fit a linear regression to all performance scores $\mathcal{P}_{k,r}$ and compute a one-sided $t$-test to decide whether to reject $H_0$ in \cref{eq:snr_x_onesided_test}. We then label the task as \emph{sensitive} to the noise-response protocol (\cref{alg:noise-calibration}) if we detect significant deterioration, and as \emph{insensitive} otherwise. Across all statistical tests we use a significance level of $\alpha=0.05$.

\section{Results}
\begin{figure}[tbp]
\centering
\includegraphics[width=\linewidth]{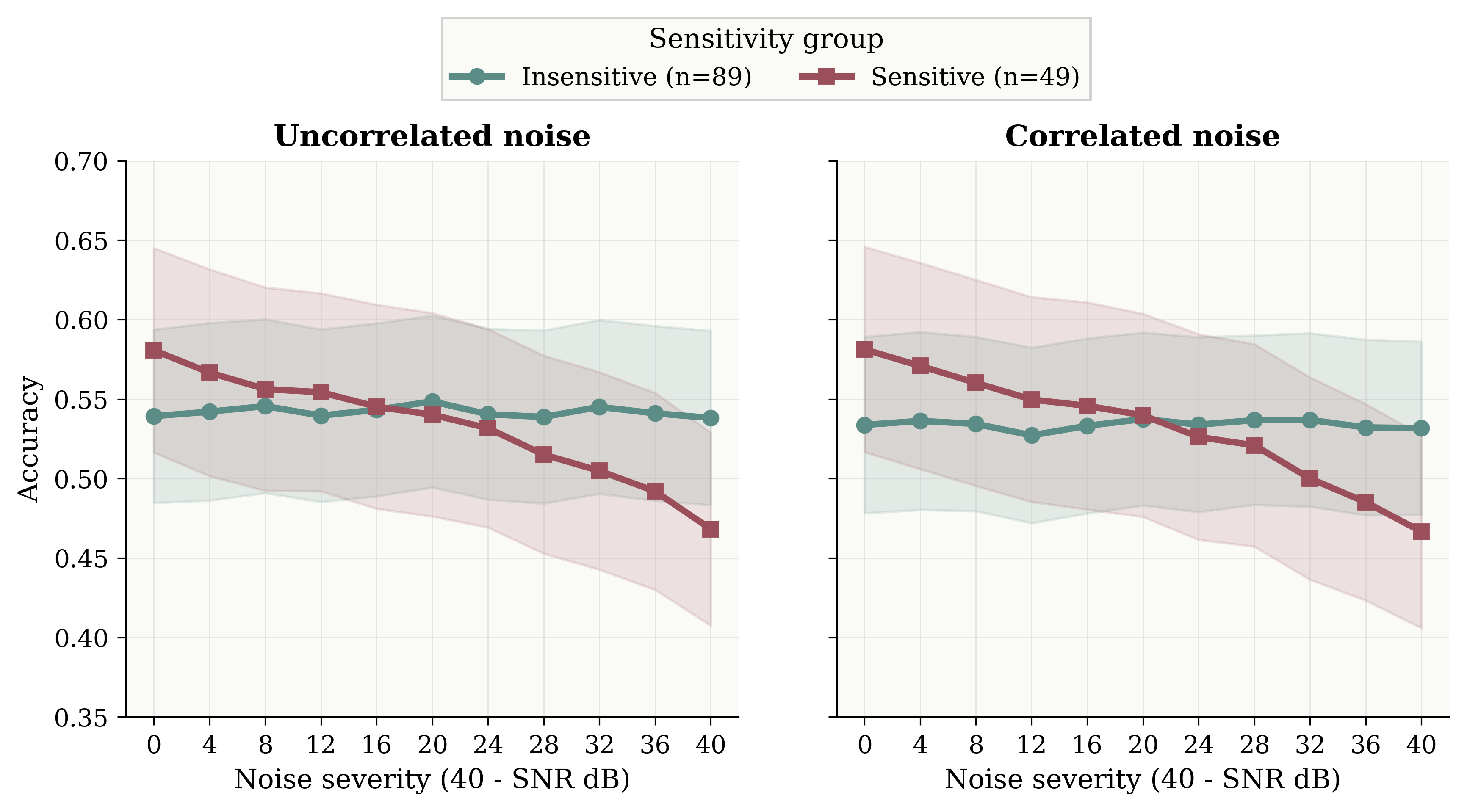}
\caption{UCI classification accuracy trajectories grouped by noise sensitivity
(Sensitive vs Insensitive). Curves are group means across datasets with 95\% CI shading,
shown separately for uncorrelated and correlated noise.}
\label{fig:uci-sensitivity-groups}
\end{figure}

\begin{figure}[tbp]
\centering
\begin{minipage}[t]{0.49\linewidth}
\centering
\includegraphics[width=\linewidth]{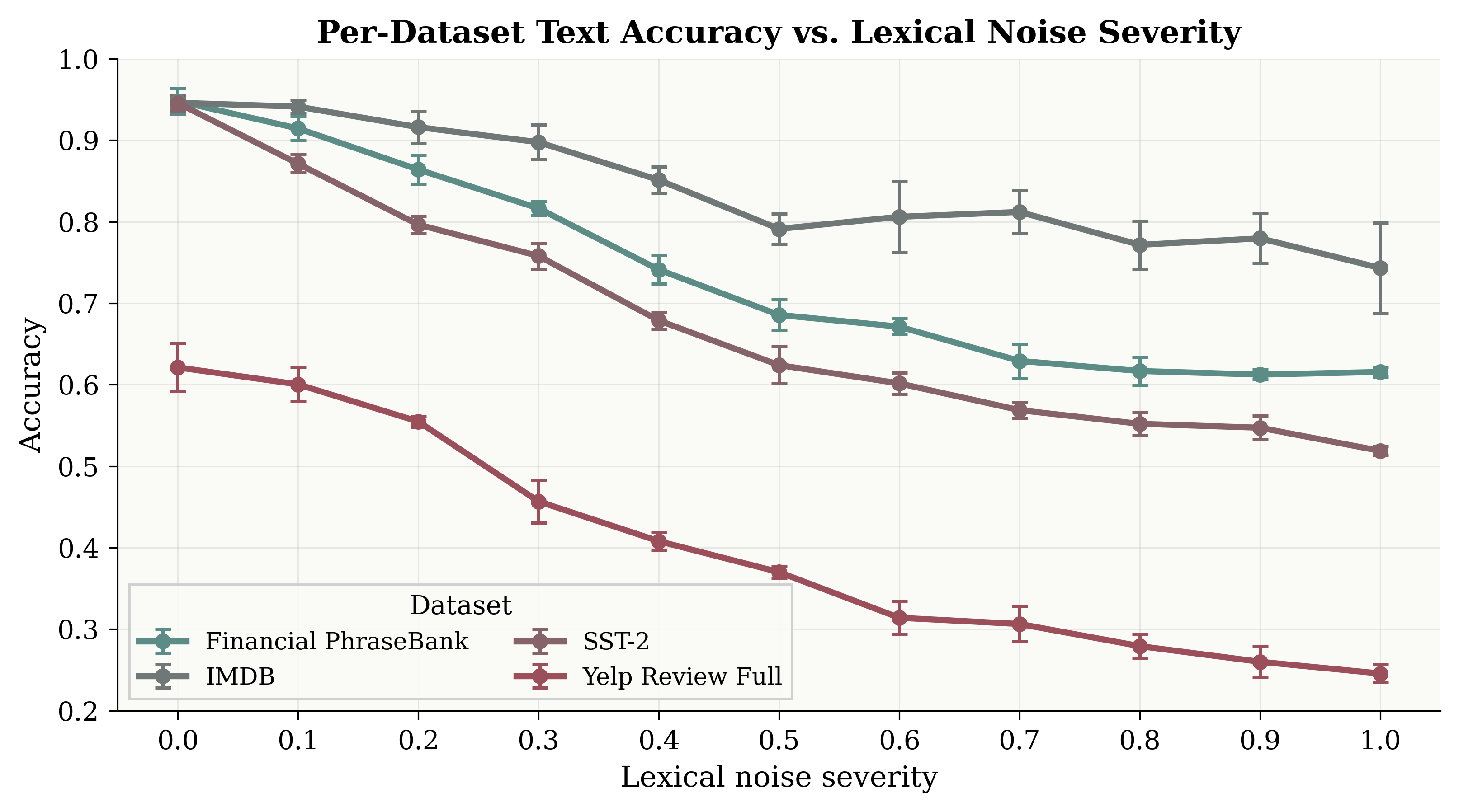}
\captionof{figure}{Per-dataset text accuracy vs lexical noise severity. Curves show means across
five repetitions, with pointwise 95\% confidence intervals.}
\label{fig:text-curves}
\end{minipage}\hfill
\begin{minipage}[t]{0.49\linewidth}
\centering
\includegraphics[width=\linewidth]{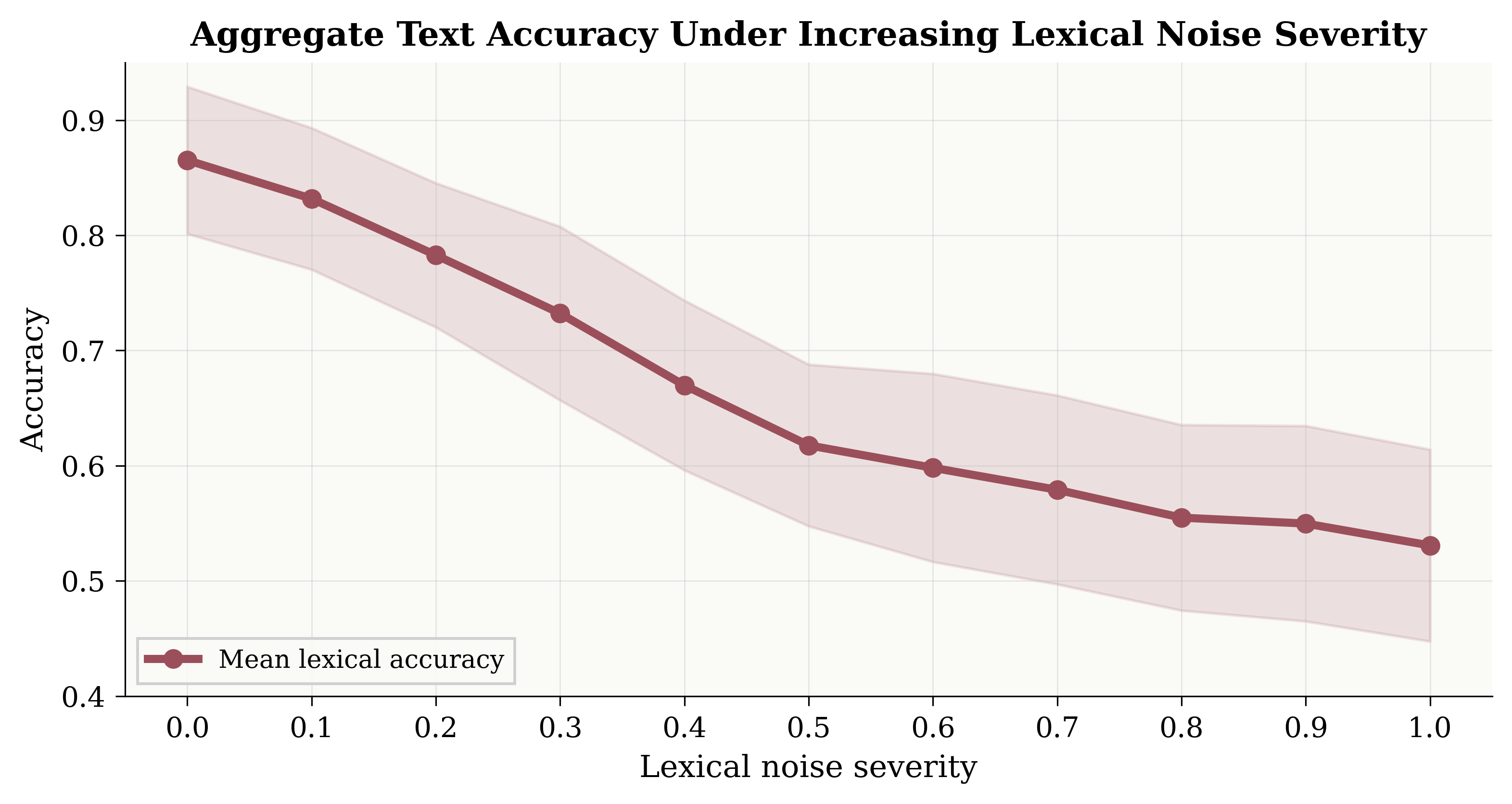}
\captionof{figure}{Aggregate text accuracy under increasing lexical noise severity.}
\label{fig:text-aggregate}
\end{minipage}
\end{figure}

We present our results in \cref{fig:uci-sensitivity-groups,fig:text-curves,fig:text-aggregate}. For the UCI tabular benchmarks, we observe a heterogeneous response to the noise interventions: 49/138 (35.5\%) classification tasks reject the null hypothesis in \cref{eq:snr_x_onesided_test} and thus exhibit sensitivity to the noise schedule. Similarly, for the regression tasks we identify 5/21 (23.8\%) tasks that react significantly to the noise schedule. One possible interpretation is that, on many tabular tasks, the LLM judge relies on shortcuts (e.g., dataset artifacts or memorized associations) rather than robust feature utilization, which may contribute to poor generalization in high-stakes use-cases. \cref{fig:uci-sensitivity-groups} shows aggregate accuracy values by sensitivity group and noise type; we see little difference between uncorrelated and correlated noise in these aggregated trajectories.

\paragraph{Tabular behavior}
While our study aimed to identify tasks where the LLM judge fails under controlled interventions, it is notable that a large fraction of UCI tasks are not measurably influenced by the tested noise patterns. This observation is consistent with recent work suggesting that LLM performance on tabular prediction remains mixed and strongly dependent on task formulation, representation, and prompting choices \citep{fang2024llmstabular,pmlr-v206-hegselmann23a,sui2024tablemeetsllm}.

\paragraph{Text-specific behavior}
We present the per-dataset trajectories in Figure~\ref{fig:text-curves}, and Figure~\ref{fig:text-aggregate}
showing the aggregate lexical trend. Lexical-noise sensitivity is high across all datasets: all four text datasets show statistically significant negative slopes under the one-sided deterioration test. IMDB appears least sensitive, while Yelp Review Full exhibits the largest deterioration. One plausible explanation is that IMDB is a binary task, where the random-guessing floor is 0.5, whereas Yelp is a 5-class sentiment task with a floor of 0.2, making large drops in accuracy more visible under severe corruption. We provide an illustration of the corruption progression in Appendix \cref{fig:appendix-imdb-lexical-progression}.

\subsection{In group vs. outgroup indications on test-set}
\begin{figure}[t]
\centering
\includegraphics[width=\linewidth]{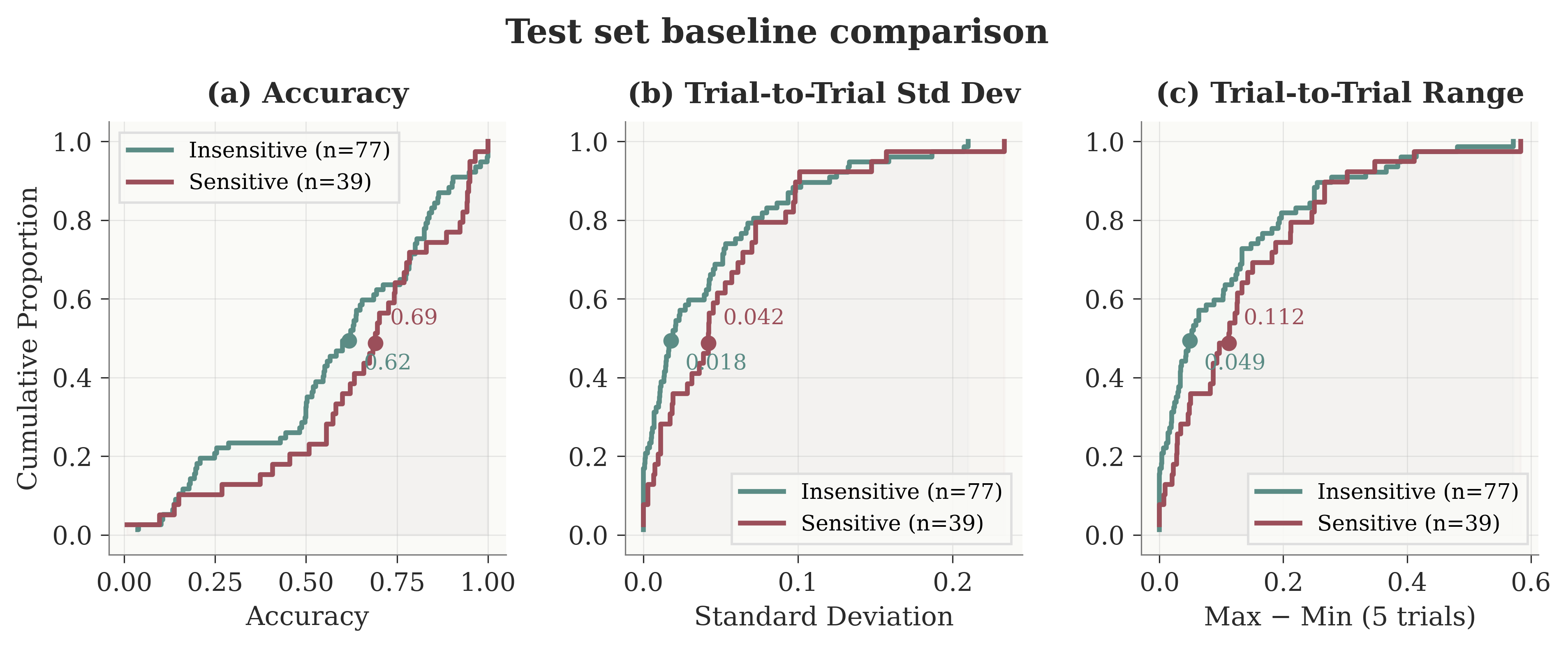}
\caption{Clean test-set baseline comparison for classification datasets (sensitive
vs insensitive): ECDFs for median accuracy, trial-to-trial standard deviation, and
trial range.}
\label{fig:testset-baselines-clean}
\end{figure}

Suppose that a dataset's sensitivity/insensitivity label is a proxy for whether an LLM judge will generalize to unseen data; how can we validate this thesis? We hypothesize that an LLM judge that proved insensitive to our noise-response protocol (\cref{alg:noise-calibration}) will generalize poorly to unseen data in the same domain. To test this, we ran an additional clean-baseline experiment (no additive noise) and compared baseline performance and variability between datasets labeled sensitive vs.\ insensitive. Due to time constraints, we limited this analysis to datasets with fewer than 1600 examples, yielding 124 datasets total (83 insensitive, 41 sensitive). As before, we run 5 repetitions per dataset and include 20 $n_{\text{context}}$ examples from the training set in the prompt. The results are summarized in \cref{fig:testset-baselines-clean} and \cref{tab:testset-baselines}. Overall, insensitive datasets tend to have lower performance; moreover, aside from baseline accuracy, dispersion measures (standard deviation, interquartile range, and trial-to-trial variation) are equal to or more than twice as large for the sensitive group. These patterns are consistent with the interpretation that insensitivity may reflect memorization or reliance on weak cues; however, higher variance in the sensitive group can also arise when predictions are closer to the decision boundary. For additional breakdowns, see Appendix \cref{tab:uci-all-slopes} and \cref{tab:uci-sensitivity-analysis-breakdown}.

\begin{table}[t]
\centering
\caption{Test-set baseline comparison (classification).
Sensitivity labels derived from training-set perturbation.
Test-set uses clean data only (5 trials/dataset).
All values are medians; CIs from 10\,000 bootstrap resamples.}
\label{tab:testset-baselines}
\small
\begin{tabular}{lrrcrc}
\toprule
Metric & Sensitive & Insensitive & Ratio
       & $\Delta$ Median & 95\% Bootstrap CI  \\
\midrule
Accuracy & 0.691 & 0.618 & 1.1$\times$ & $+0.073$ & [-0.036,\,+0.186]  \\
Std Dev & 0.0421 & 0.0178 & 2.4$\times$ & $+0.0243$ & [-0.0013,\,+0.0394]  \\
IQR & 0.0233 & 0.0116 & 2.0$\times$ & $+0.0117$ & [-0.0021,\,+0.0490]  \\
Range & 0.1121 & 0.0487 & 2.3$\times$ & $+0.0634$ & [-0.0067,\,+0.0956]  \\
\bottomrule
\end{tabular}
\end{table}

\begin{table}[tbp]
\caption{Ablation summary (mean primary metric $\pm$ std across datasets).}
\label{tab:ablation-summary}
\centering
\footnotesize
\begin{tabular}{lcccc}
\toprule
Model & 0-shot & 5-shot & 10-shot & 20-shot \\
\midrule
\textsc{gpt-4o} & 0.483 $\pm$ 0.235 & \textbf{0.509} $\pm$ 0.241 & \textbf{0.514} $\pm$ 0.240 & 0.523 $\pm$ 0.245 \\
\textsc{gpt-4o-mini} & 0.412 $\pm$ 0.210 & 0.435 $\pm$ 0.209 & 0.440 $\pm$ 0.212 & 0.443 $\pm$ 0.213 \\
\textsc{gpt-5-nano} & \textbf{0.486} $\pm$ 0.257 & 0.503 $\pm$ 0.258 & 0.509 $\pm$ 0.263 & 0.515 $\pm$ 0.266 \\
\textsc{gpt-5-mini} baseline & -- & -- & -- & \textbf{0.561} $\pm$ 0.253 \\
\bottomrule
\end{tabular}
\end{table}

\subsection{Ablations and sensitivity checks}
To guide our model choice, and in particular to assess whether few-shot examples improve performance, we ran model and few-shot ablations with \{0, 5, 10, 20\} context examples for
\textsc{gpt-4o}, \textsc{gpt-4o-mini}, and \textsc{gpt-5-nano}. As shown in \cref{tab:testset-baselines}, all three models
benefit from few-shot examples under a paired 0-shot vs.\ 20-shot comparison:
\textsc{gpt-4o} improves from 0.483 to 0.523 (+4.0\%),
\textsc{gpt-4o-mini} from 0.412 to 0.443 (+3.0\%), and
\textsc{gpt-5-nano} from 0.487 to 0.515 (+2.9 \%). At 20-shot, the
mean-performance ordering is \textsc{gpt-4o} (0.523) $>$ \textsc{gpt-5-nano}
(0.515) $>$ \textsc{gpt-4o-mini} (0.443), with a significant gap between
\textsc{gpt-4o} and \textsc{gpt-4o-mini} (+8.0\%). Notably,
\textsc{gpt-5-nano} has the strongest 0-shot baseline and shows a smaller incremental
gain from additional context than \textsc{gpt-4o}; in this ablation, \textsc{gpt-4o}
also remains competitive with the \textsc{gpt-5} family. We treat these few-shot grid
ablations as a \emph{methodological} check, to verify that (i) few-shot context helps on
these tasks and (ii) performance is meaningfully model-dependent,rather than as an
attempt to exhaustively tune the final judge configuration. Accordingly, for the main
noise-response study we fix a single strong setting (\textsc{gpt-5-mini} with 20-shot
context), aiming to operate in a regime where the judge is sufficiently capable on clean
inputs so that any observed insensitivity under noise is less likely to be an artifact
of inadequate baseline capacity.

\section{Limitations and Future Work}
\label{sec:future}
While we believe reported findings are indicative of our original thesis, namely, LLM Judges that show indifference to dramatic input changes is an indicator of an untrustworthy judge, several methodological limits remain: Our conclusions are conditional on the
chosen perturbation families, we focus on numeric-only tabular interventions and thus under-represent
datasets with substantial categorical structure, meanwhile our findings are tied to one
primary LLM and therefore do not yet establish cross-model invariance, and finally the linear
slope framework favors interpretability but can miss non-linear deterioration regimes.

An additional limitation is diagnostic depth: we identify sensitive versus
insensitive regimes but do not fully explain why insensitivity appears for specific
datasets. While we performed a simplistic analysis, providing a breakdown by categories of the UCI datasets and their findings (\cref{tab:uci-sensitivity-analysis-breakdown}), a detailed analysis that could fully exonerate potential erroneous interpretations that may not have been fully explained by the our noise protocol. This includes ablations on prompt structure, experiments where we incorporate deniability in the model output (allowing the model to decline uncertain examples) as well as variations on noise schedules not limited by numerical covariates but all covariates in the dataset. This would allow for a more comprehensive understanding of the extent by which the model is confidently uninformed or confidently informed of the problem.

\paragraph{Future directions}\mbox{}\\
Next steps are to expand the study across multiple LLM families to test whether trust
decisions are model-invariant or model-specific, extend tabular interventions to categorical and mixed-type features for broader UCI
coverage, in addition to expanding towards additional text and image datasets as well, finally we would like to add out-of-distribution plus counterfactual perturbations to separate true
causal signal dependence from potentially spurious sensitivity.

\section{Conclusion}
We introduced an intervention-based calibration protocol for LLM judges under controlled noise. The main empirical finding is a modality gap: lexical perturbations consistently degraded text performance, whereas additive Gaussian noise elicited a response in only 36\% of UCI classification tasks and 24\% of UCI regression tasks. In a post-hoc clean test-set analysis, noise-insensitive datasets tended to have lower median accuracy, while sensitive datasets showed broader trial-to-trial dispersion, suggesting that absent sensitivity may reflect brittle or weakly grounded behavior. Although preliminary, these results are consistent with noise-response sensitivity tracking whether the LLM judge is grounded in robust, task-relevant signal rather than weak or brittle cues, offering a principled starting point for improving the reliability and trust in LLM-based judges.

\bibliography{references}

\begin{thebibliography}{30}
\providecommand{\natexlab}[1]{#1}
\providecommand{\url}[1]{\texttt{#1}}
\expandafter\ifx\csname urlstyle\endcsname\relax
  \providecommand{\doi}[1]{doi: #1}\else
  \providecommand{\doi}{doi: \begingroup \urlstyle{rm}\Url}\fi

\bibitem[Asuncion \& Newman(2007)Asuncion and Newman]{Asuncion2007}
Arthur Asuncion and David Newman.
\newblock Uci machine learning repository.
\newblock University of California, Irvine, School of Information and Computer
  Sciences, 2007.
\newblock URL \url{https://archive.ics.uci.edu}.

\bibitem[Bai et~al.(2022)Bai, Kadavath, Kundu, Askell, Kernion, Jones, Chen,
  Goldie, Mirhoseini, McKinnon, Chen, Olsson, Olah, Hernandez, Drain, Ganguli,
  Li, Tran-Johnson, Perez, Kerr, Mueller, Ladish, Landau, Ndousse, Lukosuite,
  Lovitt, Sellitto, Elhage, Schiefer, Mercado, DasSarma, Lasenby, Larson,
  Ringer, Johnston, Kravec, Showk, Fort, Lanham, Telleen-Lawton, Conerly,
  Henighan, Hume, Bowman, Hatfield-Dodds, Mann, Amodei, Joseph, McCandlish,
  Brown, and Kaplan]{bai2022constitutional}
Yuntao Bai, Saurav Kadavath, Sandipan Kundu, Amanda Askell, Jackson Kernion,
  Andy Jones, Anna Chen, Anna Goldie, Azalia Mirhoseini, Cameron McKinnon,
  Carol Chen, Catherine Olsson, Christopher Olah, Danny Hernandez, Dawn Drain,
  Deep Ganguli, Dustin Li, Eli Tran-Johnson, Ethan Perez, Jamie Kerr, Jared
  Mueller, Jeffrey Ladish, Joshua Landau, Kamal Ndousse, Kamile Lukosuite,
  Liane Lovitt, Michael Sellitto, Nelson Elhage, Nicholas Schiefer, Noemi
  Mercado, Nova DasSarma, Robert Lasenby, Robin Larson, Sam Ringer, Scott
  Johnston, Shauna Kravec, Sheer~El Showk, Stanislav Fort, Tamera Lanham,
  Timothy Telleen-Lawton, Tom Conerly, Tom Henighan, Tristan Hume, Samuel~R.
  Bowman, Zac Hatfield-Dodds, Ben Mann, Dario Amodei, Nicholas Joseph, Sam
  McCandlish, Tom Brown, and Jared Kaplan.
\newblock Constitutional ai: Harmlessness from ai feedback, 2022.
\newblock URL \url{https://arxiv.org/abs/2212.08073}.

\bibitem[Belinkov \& Bisk(2018)Belinkov and Bisk]{belinkov2018synthetic}
Yonatan Belinkov and Yonatan Bisk.
\newblock Synthetic and natural noise both break neural machine translation.
\newblock In \emph{Proceedings of ICLR}, 2018.
\newblock URL \url{https://openreview.net/forum?id=BJ8vJebC-}.

\bibitem[Cortez et~al.(2009)Cortez, Cerdeira, Almeida, Matos, and
  Reis]{cortez2009wine}
Paulo Cortez, Ant{\'o}nio Cerdeira, Fernando Almeida, Telmo Matos, and Jos{\'e}
  Reis.
\newblock Modeling wine preferences by data mining from physicochemical
  properties.
\newblock \emph{Decision Support Systems}, 47\penalty0 (4):\penalty0 547--553,
  2009.
\newblock \doi{10.1016/j.dss.2009.05.016}.

\bibitem[DeMillo et~al.(1978)DeMillo, Lipton, and Sayward]{demillo1978hints}
Richard~A. DeMillo, Richard~J. Lipton, and Frederick~G. Sayward.
\newblock Hints on test data selection: Help for the practicing programmer.
\newblock \emph{Computer}, 11\penalty0 (4):\penalty0 34--41, 1978.
\newblock \doi{10.1109/C-M.1978.218136}.

\bibitem[Fang et~al.(2024)Fang, Xu, Tan, Zhang, Hu, Qi, Nickleach, Socolinsky,
  Sengamedu, and Faloutsos]{fang2024llmstabular}
Xi~Fang, Weijie Xu, Fiona~Anting Tan, Jiani Zhang, Ziqing Hu, Yanjun Qi, Scott
  Nickleach, Diego Socolinsky, Srinivasan~H. Sengamedu, and Christos Faloutsos.
\newblock Large language models (llms) on tabular data: Prediction, generation,
  and understanding -- a survey.
\newblock \emph{CoRR}, abs/2402.17944, 2024.
\newblock \doi{10.48550/arXiv.2402.17944}.
\newblock URL \url{https://doi.org/10.48550/arXiv.2402.17944}.

\bibitem[Fisher(1936)]{fisher1936iris}
Ronald~A. Fisher.
\newblock The use of multiple measurements in taxonomic problems.
\newblock \emph{Annals of Eugenics}, 7\penalty0 (2):\penalty0 179--188, 1936.

\bibitem[Gebru et~al.(2021)Gebru, Morgenstern, Vecchione, Vaughan, Wallach,
  III, and Crawford]{gebru2021datasheetsdatasets}
Timnit Gebru, Jamie Morgenstern, Briana Vecchione, Jennifer~Wortman Vaughan,
  Hanna Wallach, Hal~Daum\'{e} III, and Kate Crawford.
\newblock Datasheets for datasets.
\newblock \emph{Commun. ACM}, 64\penalty0 (12):\penalty0 86–92, November
  2021.
\newblock ISSN 0001-0782.
\newblock \doi{10.1145/3458723}.
\newblock URL \url{https://doi.org/10.1145/3458723}.

\bibitem[Hegselmann et~al.(2023)Hegselmann, Buendia, Lang, Agrawal, Jiang, and
  Sontag]{pmlr-v206-hegselmann23a}
Stefan Hegselmann, Alejandro Buendia, Hunter Lang, Monica Agrawal, Xiaoyi
  Jiang, and David Sontag.
\newblock Tabllm: Few-shot classification of tabular data with large language
  models.
\newblock In \emph{Proceedings of The 26th International Conference on
  Artificial Intelligence and Statistics}, volume 206 of \emph{Proceedings of
  Machine Learning Research}, pp.\  5549--5581. PMLR, 2023.
\newblock URL \url{https://proceedings.mlr.press/v206/hegselmann23a.html}.

\bibitem[Huang et~al.(2025)Huang, Yu, Ma, Zhong, Feng, Wang, Chen, Peng, Feng,
  Qin, and Liu]{huang2024surveyhallucinationlargelanguage}
Lei Huang, Weijiang Yu, Weitao Ma, Weihong Zhong, Zhangyin Feng, Haotian Wang,
  Qianglong Chen, Weihua Peng, Xiaocheng Feng, Bing Qin, and Ting Liu.
\newblock A survey on hallucination in large language models: Principles,
  taxonomy, challenges, and open questions.
\newblock \emph{ACM Transactions on Information Systems}, 43\penalty0
  (2):\penalty0 1–55, January 2025.
\newblock ISSN 1558-2868.
\newblock \doi{10.1145/3703155}.
\newblock URL \url{http://dx.doi.org/10.1145/3703155}.

\bibitem[Jia \& Harman(2011)Jia and Harman]{jia2011analysis}
Yue Jia and Mark Harman.
\newblock An analysis and survey of the development of mutation testing.
\newblock \emph{IEEE Transactions on Software Engineering}, 37\penalty0
  (5):\penalty0 649--678, 2011.
\newblock \doi{10.1109/TSE.2010.62}.

\bibitem[Lee et~al.(2026)Lee, Zeng, Jeong, yong Sohn, and Lee]{lee2025llmjudge}
Chungpa Lee, Thomas Zeng, Jongwon Jeong, Jy~yong Sohn, and Kangwook Lee.
\newblock How to correctly report llm-as-a-judge evaluations, 2026.
\newblock URL \url{https://arxiv.org/abs/2511.21140}.

\bibitem[Lin et~al.(2022)Lin, Hilton, and
  Evans]{lin2022truthfulqameasuringmodelsmimic}
Stephanie Lin, Jacob Hilton, and Owain Evans.
\newblock {T}ruthful{QA}: Measuring how models mimic human falsehoods.
\newblock In Smaranda Muresan, Preslav Nakov, and Aline Villavicencio (eds.),
  \emph{Proceedings of the 60th Annual Meeting of the Association for
  Computational Linguistics (Volume 1: Long Papers)}, pp.\  3214--3252, Dublin,
  Ireland, May 2022. Association for Computational Linguistics.
\newblock \doi{10.18653/v1/2022.acl-long.229}.
\newblock URL \url{https://aclanthology.org/2022.acl-long.229/}.

\bibitem[Liu et~al.(2023)Liu, Iter, Xu, Wang, Xu, and
  Zhu]{liu2023gevalnlgevaluationusing}
Yang Liu, Dan Iter, Yichong Xu, Shuohang Wang, Ruochen Xu, and Chenguang Zhu.
\newblock G-eval: {NLG} evaluation using {GPT}-4 with better human alignment,
  2023.
\newblock URL \url{https://arxiv.org/abs/2303.16634}.

\bibitem[Maas et~al.(2011)Maas, Daly, Pham, Huang, Ng, and Potts]{maas2011imdb}
Andrew~L. Maas, Raymond~E. Daly, Peter~T. Pham, Dan Huang, Andrew~Y. Ng, and
  Christopher Potts.
\newblock Learning word vectors for sentiment analysis.
\newblock In \emph{Proceedings of ACL: Human Language Technologies}, pp.\
  142--150, 2011.
\newblock URL \url{https://aclanthology.org/P11-1015/}.

\bibitem[Malo et~al.(2014)Malo, Sinha, Korhonen, Wallenius, and
  Takala]{malo2014gooddebt}
Pekka Malo, Ankur Sinha, Pekka Korhonen, Jyrki Wallenius, and Pyry Takala.
\newblock Good debt or bad debt: Detecting semantic orientations in economic
  texts.
\newblock \emph{Journal of the Association for Information Science and
  Technology}, 65\penalty0 (4):\penalty0 782--796, 2014.
\newblock \doi{10.1002/asi.23062}.

\bibitem[Mirzadeh et~al.(2025)Mirzadeh, Alizadeh, Shahrokhi, Tuzel, Bengio, and
  Farajtabar]{mirzadeh2024gsmsymbolic}
Iman Mirzadeh, Keivan Alizadeh, Hooman Shahrokhi, Oncel Tuzel, Samy Bengio, and
  Mehrdad Farajtabar.
\newblock Gsm-symbolic: Understanding the limitations of mathematical reasoning
  in large language models, 2025.
\newblock URL \url{https://arxiv.org/abs/2410.05229}.

\bibitem[Ouyang et~al.(2022)Ouyang, Wu, Jiang, Almeida, Wainwright, Mishkin,
  Zhang, Agarwal, Slama, Ray, Schulman, Hilton, Kelton, Miller, Simens, Askell,
  Welinder, Christiano, Leike, and Lowe]{ouyang2022training}
Long Ouyang, Jeff Wu, Xu~Jiang, Diogo Almeida, Carroll~L. Wainwright, Pamela
  Mishkin, Chong Zhang, Sandhini Agarwal, Katarina Slama, Alex Ray, John
  Schulman, Jacob Hilton, Fraser Kelton, Luke Miller, Maddie Simens, Amanda
  Askell, Peter Welinder, Paul Christiano, Jan Leike, and Ryan Lowe.
\newblock Training language models to follow instructions with human feedback,
  2022.
\newblock URL \url{https://arxiv.org/abs/2203.02155}.

\bibitem[Pearl(2009)]{pearl2009causality}
Judea Pearl.
\newblock \emph{Causality}.
\newblock Cambridge University Press, 2009.

\bibitem[Peters et~al.(2017)Peters, Janzing, and Scholkopf]{peters2017elements}
Jonas Peters, Dominik Janzing, and Bernhard Scholkopf.
\newblock \emph{Elements of Causal Inference}.
\newblock MIT Press, 2017.

\bibitem[Pruthi et~al.(2019)Pruthi, Dhingra, and Lipton]{pruthi2019combating}
Danish Pruthi, Bhuwan Dhingra, and Zachary~C. Lipton.
\newblock Combating adversarial misspellings with robust word recognition.
\newblock In \emph{Proceedings of ACL}, pp.\  5582--5591, 2019.
\newblock URL \url{https://aclanthology.org/P19-1561/}.

\bibitem[Seber \& Lee(2012)Seber and Lee]{seber2012linear}
George A.~F. Seber and Alan~J. Lee.
\newblock \emph{Linear Regression Analysis}.
\newblock John Wiley \& Sons, 2nd edition, 2012.
\newblock \doi{10.1002/9780471722199}.

\bibitem[Seedat et~al.(2024)Seedat, Huynh, van Breugel, and van~der
  Schaar]{seedat2024curatedllmsynergyllms}
Nabeel Seedat, Nicolas Huynh, Boris van Breugel, and Mihaela van~der Schaar.
\newblock Curated llm: Synergy of llms and data curation for tabular
  augmentation in low-data regimes, 2024.
\newblock URL \url{https://arxiv.org/abs/2312.12112}.

\bibitem[Shi et~al.(2023)Shi, Chen, Misra, Scales, Dohan, Chi, Schärli, and
  Zhou]{shi2023distracted}
Freda Shi, Xinyun Chen, Kanishka Misra, Nathan Scales, David Dohan, Ed~Chi,
  Nathanael Schärli, and Denny Zhou.
\newblock Large language models can be easily distracted by irrelevant context,
  2023.
\newblock URL \url{https://arxiv.org/abs/2302.00093}.

\bibitem[Snow et~al.(2008)Snow, O'Connor, Jurafsky, and
  Ng]{snow-etal-2008-cheap}
Rion Snow, Brendan O'Connor, Daniel Jurafsky, and Andrew Ng.
\newblock Cheap and fast -- but is it good? evaluating non-expert annotations
  for natural language tasks.
\newblock In \emph{Proceedings of EMNLP}, pp.\  254--263, 2008.
\newblock URL \url{https://aclanthology.org/D08-1027/}.

\bibitem[Socher et~al.(2013)Socher, Perelygin, Wu, Chuang, Manning, Ng, and
  Potts]{socher2013recursive}
Richard Socher, Alex Perelygin, Jean Wu, Jason Chuang, Christopher~D. Manning,
  Andrew Ng, and Christopher Potts.
\newblock Recursive deep models for semantic compositionality over a sentiment
  treebank.
\newblock In \emph{Proceedings of EMNLP}, pp.\  1631--1642, 2013.
\newblock URL \url{https://aclanthology.org/D13-1170/}.

\bibitem[Sui et~al.(2024)Sui, Zhou, Zhou, Han, and Zhang]{sui2024tablemeetsllm}
Yuan Sui, Mengyu Zhou, Mingjie Zhou, Shi Han, and Dongmei Zhang.
\newblock Table meets llm: Can large language models understand structured
  table data? a benchmark and empirical study.
\newblock In \emph{Proceedings of the 17th ACM International Conference on Web
  Search and Data Mining}, pp.\  645--654, 2024.
\newblock \doi{10.1145/3616855.3635752}.
\newblock URL \url{https://doi.org/10.1145/3616855.3635752}.

\bibitem[Wang et~al.(2019)Wang, Singh, Michael, Hill, Levy, and
  Bowman]{wang2019glue}
Alex Wang, Amanpreet Singh, Julian Michael, Felix Hill, Omer Levy, and
  Samuel~R. Bowman.
\newblock {GLUE}: A multi-task benchmark and analysis platform for natural
  language understanding.
\newblock In \emph{Proceedings of ICLR}, 2019.
\newblock URL \url{https://openreview.net/forum?id=rJ4km2R5t7}.

\bibitem[Zhang et~al.(2015)Zhang, Zhao, and LeCun]{zhang2015character}
Xiang Zhang, Junbo Zhao, and Yann LeCun.
\newblock Character-level convolutional networks for text classification.
\newblock In \emph{Proceedings of NeurIPS}, 2015.
\newblock URL
  \url{https://proceedings.neurips.cc/paper/2015/hash/250cf8b51c773f3f8dc8b4be867a9a02-Abstract.html}.

\bibitem[Zheng et~al.(2023)Zheng, Chiang, Sheng, Zhuang, Wu, Zhuang, Lin, Li,
  Li, Xing, Zhang, Gonzalez, and
  Stoica]{zheng2023judgingllmasajudgemtbenchchatbot}
Lianmin Zheng, Wei-Lin Chiang, Ying Sheng, Siyuan Zhuang, Zhanghao Wu, Yonghao
  Zhuang, Zi~Lin, Zhuohan Li, Dacheng Li, Eric~P. Xing, Hao Zhang, Joseph~E.
  Gonzalez, and Ion Stoica.
\newblock Judging {LLM}-as-a-judge with {MT}-bench and chatbot arena, 2023.
\newblock URL \url{https://arxiv.org/abs/2306.05685}.

\end{thebibliography}
\bibliographystyle{iclr2026_conference}
\clearpage
\appendix
\section{Additional Results}
\subsection{Per-dataset text lexical slopes}
Table~\ref{tab:text-slopes-report} lists per-dataset lexical slope estimates for text
experiments.

\begin{table}[tbp]
\caption{Lexical slope estimates ($\beta_1$), one-sided $p$-values, and $H_0$ decisions.}
\label{tab:text-slopes-report}
\centering
\begin{tabular}{lccc}
\toprule
Dataset & Lexical $\beta_1$ & One-sided $p$ & Decision \\
\midrule
Financial PhraseBank & -0.3687 & $<10^{-10}$ & Reject $H_0$ \\
IMDB & -0.2100 & $<10^{-10}$ & Reject $H_0$ \\
SST-2 & -0.4199 & $<10^{-10}$ & Reject $H_0$ \\
Yelp Review Full & -0.4056 & $<10^{-10}$ & Reject $H_0$ \\
\bottomrule
\end{tabular}
\end{table}

\subsection{All UCI per-dataset slopes}
Table~\ref{tab:uci-all-slopes} reports UCI per-dataset slope estimates under
uncorrelated and correlated noise, for both classification (accuracy) and regression
($R^2$). This table covers the full set of datasets with complete per-dataset
trajectories in the experiments reported in this paper.

\begingroup
\scriptsize
\setlength{\tabcolsep}{3.5pt}
\begin{longtable}{l p{4.0cm} r r r r l}
\caption{Per-dataset UCI slope estimates ($\beta_1$) from linear regression of primary metric on noise severity ($40-\mathrm{SNR}_{\mathrm{dB}}$), with one-sided $p$-values for $H_1:\beta_1<0$. Buckets are derived using the sensitivity/insensitivity deterioration protocol defined in \cref{alg:noise-calibration}. Classification (n=138): Insensitive=89, Sensitive (both)=32, Sensitive (corr-only)=11, Sensitive (uncorr-only)=6. Regression (n=21): Insensitive=16, Sensitive (both)=3, Sensitive (corr-only)=0, Sensitive (uncorr-only)=2.}\label{tab:uci-all-slopes} \\
\toprule
Task & Dataset & $\beta_1$ (Uncorr.) & $p$ (Uncorr.) & $\beta_1$ (Corr.) & $p$ (Corr.) & Bucket \\
\midrule
\endfirsthead
\multicolumn{7}{l}{\textit{Table \thetable\ continued}} \\
\toprule
Task & Dataset & $\beta_1$ (Uncorr.) & $p$ (Uncorr.) & $\beta_1$ (Corr.) & $p$ (Corr.) & Bucket \\
\midrule
\endhead
\midrule
\multicolumn{7}{r}{\textit{Continued on next page}} \\
\endfoot
\bottomrule
\endlastfoot
Classification & \texttt{abalone} & -0.0002 & 0.1896 & -0.0007 & 0.0119 & Sensitive (corr-only) \\
Classification & \texttt{acute\_\allowbreak{}inflammations} & -0.0002 & 0.4039 & -0.0000 & 0.4726 & Insensitive \\
Classification & \texttt{adult} & +0.0008 & 0.8151 & -0.0012 & 0.1322 & Insensitive \\
Classification & \texttt{android\_\allowbreak{}permissions} & +0.0000 & 0.7318 & -0.0001 & 0.2699 & Insensitive \\
Classification & \texttt{annealing} & +0.0009 & 0.6829 & +0.0023 & 0.8607 & Insensitive \\
Classification & \texttt{auction\_\allowbreak{}verification} & -0.0000 & 0.1738 & -0.0000 & 0.1047 & Insensitive \\
Classification & \texttt{audiology} & +0.0002 & 0.6896 & -0.0002 & 0.2878 & Insensitive \\
Classification & \texttt{autism\_\allowbreak{}screening\_\allowbreak{}adult} & -0.0002 & 0.2608 & -0.0007 & 0.0130 & Sensitive (corr-only) \\
Classification & \texttt{autism\_\allowbreak{}screening\_\allowbreak{}children} & -0.0055 & $< 1e^{-4}$ & -0.0049 & $< 1e^{-4}$ & Sensitive (both) \\
Classification & \texttt{automobile} & -- & -- & -- & -- & Insensitive \\
Classification & \texttt{balance\_\allowbreak{}scale} & -0.0087 & $< 1e^{-4}$ & -0.0076 & $< 1e^{-4}$ & Sensitive (both) \\
Classification & \texttt{balloons} & +0.0032 & 0.9774 & -0.0014 & 0.1878 & Insensitive \\
Classification & \texttt{bank\_\allowbreak{}marketing} & -0.0001 & 0.2688 & +0.0000 & 0.5582 & Insensitive \\
Classification & \texttt{banknote\_\allowbreak{}authentication} & -0.0009 & 0.0874 & -0.0004 & 0.2470 & Insensitive \\
Classification & \texttt{blood\_\allowbreak{}transfusion} & -0.0001 & 0.4184 & -0.0007 & 0.0709 & Insensitive \\
Classification & \texttt{bone\_\allowbreak{}marrow\_\allowbreak{}transplant} & -0.0002 & 0.2957 & -- & -- & Insensitive \\
Classification & \texttt{breast\_\allowbreak{}cancer} & -0.0006 & 0.0221 & -0.0005 & 0.0484 & Sensitive (both) \\
Classification & \texttt{breast\_\allowbreak{}cancer\_\allowbreak{}coimbra} & -0.0002 & 0.3720 & +0.0010 & 0.9478 & Insensitive \\
Classification & \texttt{breast\_\allowbreak{}cancer\_\allowbreak{}wisconsin\_\allowbreak{}diagnostic} & -0.0010 & 0.0035 & -0.0007 & 0.0291 & Sensitive (uncorr-only) \\
Classification & \texttt{breast\_\allowbreak{}cancer\_\allowbreak{}wisconsin\_\allowbreak{}original} & -0.0076 & $< 1e^{-4}$ & -0.0066 & $< 1e^{-4}$ & Sensitive (both) \\
Classification & \texttt{breast\_\allowbreak{}cancer\_\allowbreak{}wisconsin\_\allowbreak{}prognostic} & +0.0004 & 0.9005 & +0.0000 & 0.5134 & Sensitive (corr-only) \\
Classification & \texttt{car\_\allowbreak{}evaluation} & -0.0005 & 0.3019 & +0.0001 & 0.5493 & Insensitive \\
Classification & \texttt{cardiotocography} & +0.0000 & 0.5000 & +0.0001 & 0.8262 & Insensitive \\
Classification & \texttt{cdc\_\allowbreak{}diabetes\_\allowbreak{}indicators} & -0.0003 & 0.3448 & -0.0001 & 0.4114 & Insensitive \\
Classification & \texttt{census\_\allowbreak{}income} & -0.0024 & 0.0038 & -0.0002 & 0.4092 & Sensitive (uncorr-only) \\
Classification & \texttt{cervical\_\allowbreak{}cancer\_\allowbreak{}behavior\_\allowbreak{}risk} & -0.0018 & 0.0534 & -0.0015 & 0.0940 & Sensitive (both) \\
Classification & \texttt{chess\_\allowbreak{}king\_\allowbreak{}rook\_\allowbreak{}vs\_\allowbreak{}king} & -0.0001 & 0.1147 & +0.0000 & 0.6009 & Insensitive \\
Classification & \texttt{chess\_\allowbreak{}king\_\allowbreak{}rook\_\allowbreak{}vs\_\allowbreak{}king\_\allowbreak{}pawn} & +0.0005 & 0.9968 & -0.0002 & 0.0851 & Insensitive \\
Classification & \texttt{chronic\_\allowbreak{}kidney\_\allowbreak{}disease} & -0.0035 & $< 1e^{-4}$ & -0.0037 & $< 1e^{-4}$ & Sensitive (both) \\
Classification & \texttt{cirrhosis\_\allowbreak{}survival} & -0.0036 & $< 1e^{-4}$ & -0.0033 & $< 1e^{-4}$ & Sensitive (both) \\
Classification & \texttt{ckd\_\allowbreak{}risk\_\allowbreak{}factors} & +0.0004 & 0.6566 & -0.0007 & 0.2311 & Insensitive \\
Classification & \texttt{connect\_\allowbreak{}4} & -0.0001 & 0.4264 & +0.0010 & 0.8688 & Insensitive \\
Classification & \texttt{contraceptive\_\allowbreak{}method\_\allowbreak{}choice} & +0.0006 & 0.8752 & +0.0011 & 0.9872 & Insensitive \\
Classification & \texttt{covertype} & +0.0002 & 0.7323 & -0.0003 & 0.2363 & Insensitive \\
Classification & \texttt{credit\_\allowbreak{}approval} & -0.0009 & 0.0010 & -0.0008 & 0.0099 & Sensitive (both) \\
Classification & \texttt{cylinder\_\allowbreak{}bands} & -0.0000 & 0.1044 & +0.0000 & 0.6225 & Insensitive \\
Classification & \texttt{default\_\allowbreak{}credit\_\allowbreak{}card} & -0.0020 & 0.0818 & -0.0038 & 0.0061 & Sensitive (corr-only) \\
Classification & \texttt{dermatology} & +0.0002 & 0.6595 & -0.0001 & 0.3545 & Insensitive \\
Classification & \texttt{diabetic\_\allowbreak{}retinopathy} & +0.0001 & 0.6208 & +0.0000 & 0.5718 & Insensitive \\
Classification & \texttt{dota2\_\allowbreak{}games} & -0.0000 & 0.4447 & -0.0000 & 0.4794 & Insensitive \\
Classification & \texttt{drug\_\allowbreak{}consumption} & -0.0001 & 0.2661 & +0.0000 & 0.5000 & Insensitive \\
Classification & \texttt{dry\_\allowbreak{}bean} & -0.0043 & $< 1e^{-4}$ & -0.0016 & 0.0174 & Sensitive (both) \\
Classification & \texttt{early\_\allowbreak{}stage\_\allowbreak{}diabetes} & +0.0003 & 0.6591 & -0.0010 & 0.0622 & Insensitive \\
Classification & \texttt{ecoli} & -0.0039 & $< 1e^{-4}$ & -0.0028 & $< 1e^{-4}$ & Sensitive (both) \\
Classification & \texttt{eeg\_\allowbreak{}eye\_\allowbreak{}state} & +0.0000 & 0.5000 & +0.0000 & 0.5000 & Insensitive \\
Classification & \texttt{electrical\_\allowbreak{}grid\_\allowbreak{}stability} & -0.0000 & 0.4468 & -0.0000 & 0.2516 & Insensitive \\
Classification & \texttt{fertility} & -0.0003 & 0.3350 & -0.0009 & 0.0803 & Insensitive \\
Classification & \texttt{flags} & +0.0003 & 0.8951 & -0.0000 & 0.4622 & Insensitive \\
Classification & \texttt{garment\_\allowbreak{}productivity} & -0.0001 & 0.0089 & -0.0001 & 0.0020 & Sensitive (both) \\
Classification & \texttt{gender\_\allowbreak{}by\_\allowbreak{}name} & +0.0000 & 0.5251 & -0.0001 & 0.2565 & Insensitive \\
Classification & \texttt{glass\_\allowbreak{}identification} & -0.0036 & $< 1e^{-4}$ & -0.0032 & $< 1e^{-4}$ & Sensitive (both) \\
Classification & \texttt{haberman\_\allowbreak{}survival} & -0.0018 & $< 1e^{-4}$ & -0.0024 & $< 1e^{-4}$ & Sensitive (both) \\
Classification & \texttt{hayes\_\allowbreak{}roth} & -0.0049 & $< 1e^{-4}$ & -0.0055 & $< 1e^{-4}$ & Sensitive (both) \\
Classification & \texttt{hcv\_\allowbreak{}data} & -0.0045 & $< 1e^{-4}$ & -0.0045 & $< 1e^{-4}$ & Sensitive (both) \\
Classification & \texttt{heart\_\allowbreak{}disease} & -0.0047 & $< 1e^{-4}$ & -0.0042 & $< 1e^{-4}$ & Sensitive (both) \\
Classification & \texttt{heart\_\allowbreak{}failure\_\allowbreak{}clinical\_\allowbreak{}records} & -0.0045 & $< 1e^{-4}$ & -0.0041 & $< 1e^{-4}$ & Sensitive (both) \\
Classification & \texttt{hepatitis} & -0.0004 & 0.0301 & -0.0005 & 0.0170 & Sensitive (corr-only) \\
Classification & \texttt{hepatitis\_\allowbreak{}c\_\allowbreak{}virus} & -0.0001 & 0.0843 & +0.0000 & 0.7298 & Insensitive \\
Classification & \texttt{higher\_\allowbreak{}education\_\allowbreak{}performance} & +0.0002 & 0.6734 & -0.0007 & 0.0295 & Insensitive \\
Classification & \texttt{horse\_\allowbreak{}colic} & -0.0015 & 0.0199 & -- & -- & Sensitive (uncorr-only) \\
Classification & \texttt{htru2} & -0.0008 & 0.0009 & -0.0012 & 0.0006 & Sensitive (both) \\
Classification & \texttt{ilpd} & +0.0019 & 1.0000 & +0.0017 & 1.0000 & Insensitive \\
Classification & \texttt{image\_\allowbreak{}segmentation} & -0.0025 & 0.0507 & -0.0044 & 0.0092 & Sensitive (both) \\
Classification & \texttt{in\_\allowbreak{}vehicle\_\allowbreak{}coupon} & +0.0000 & 0.5475 & +0.0002 & 0.7485 & Insensitive \\
Classification & \texttt{ionosphere} & -0.0001 & 0.3138 & -0.0002 & 0.0337 & Sensitive (corr-only) \\
Classification & \texttt{iranian\_\allowbreak{}churn} & +0.0003 & 0.8269 & -0.0004 & 0.1692 & Insensitive \\
Classification & \texttt{iris} & -0.0092 & $< 1e^{-4}$ & -0.0084 & $< 1e^{-4}$ & Sensitive (both) \\
Classification & \texttt{isolet} & -0.0000 & 0.2703 & -0.0001 & 0.0344 & Insensitive \\
Classification & \texttt{japanese\_\allowbreak{}credit\_\allowbreak{}screening} & -0.0001 & 0.0504 & -0.0000 & 0.1616 & Insensitive \\
Classification & \texttt{lenses} & -0.0012 & 0.1729 & -0.0005 & 0.3867 & Insensitive \\
Classification & \texttt{letter\_\allowbreak{}recognition} & +0.0001 & 0.8953 & +0.0001 & 0.7859 & Insensitive \\
Classification & \texttt{liver\_\allowbreak{}disorders} & -0.0010 & 0.0089 & -0.0011 & 0.0328 & Sensitive (both) \\
Classification & \texttt{lung\_\allowbreak{}cancer} & -0.0010 & 0.0368 & -0.0008 & 0.1179 & Insensitive \\
Classification & \texttt{lymphography} & -0.0003 & 0.3359 & -0.0009 & 0.1317 & Insensitive \\
Classification & \texttt{magic\_\allowbreak{}gamma\_\allowbreak{}telescope} & +0.0004 & 0.8297 & -0.0003 & 0.2583 & Insensitive \\
Classification & \texttt{mammographic\_\allowbreak{}mass} & -0.0048 & $< 1e^{-4}$ & -0.0036 & $< 1e^{-4}$ & Sensitive (both) \\
Classification & \texttt{maternal\_\allowbreak{}health\_\allowbreak{}risk} & -0.0030 & $< 1e^{-4}$ & -0.0032 & $< 1e^{-4}$ & Sensitive (both) \\
Classification & \texttt{molecular\_\allowbreak{}biology\_\allowbreak{}splice} & +0.0000 & 0.5000 & +0.0000 & 0.5000 & Insensitive \\
Classification & \texttt{monks\_\allowbreak{}problems} & -0.0049 & $< 1e^{-4}$ & -0.0050 & $< 1e^{-4}$ & Sensitive (both) \\
Classification & \texttt{mushroom} & +0.0004 & 0.7615 & -0.0005 & 0.1523 & Insensitive \\
Classification & \texttt{musk\_\allowbreak{}v1} & -0.0006 & 0.2023 & +0.0009 & 0.9304 & Insensitive \\
Classification & \texttt{musk\_\allowbreak{}v2} & +0.0005 & 0.8262 & +0.0008 & 0.8238 & Insensitive \\
Classification & \texttt{myocardial\_\allowbreak{}infarction} & +0.0000 & 0.5767 & -- & -- & Insensitive \\
Classification & \texttt{nursery} & +0.0017 & 0.9731 & -0.0015 & 0.0244 & Sensitive (corr-only) \\
Classification & \texttt{obesity\_\allowbreak{}levels} & -0.0057 & 0.0007 & -0.0065 & 0.0001 & Sensitive (both) \\
Classification & \texttt{occupancy\_\allowbreak{}detection} & -0.0001 & 0.4307 & -0.0003 & 0.2215 & Insensitive \\
Classification & \texttt{online\_\allowbreak{}shoppers\_\allowbreak{}intention} & -0.0008 & 0.1209 & -0.0005 & 0.1880 & Insensitive \\
Classification & \texttt{optical\_\allowbreak{}recognition\_\allowbreak{}digits} & -0.0000 & 0.2442 & -0.0002 & 0.0024 & Sensitive (corr-only) \\
Classification & \texttt{ozone\_\allowbreak{}level\_\allowbreak{}detection} & +0.0023 & 0.9887 & -0.0001 & 0.4521 & Insensitive \\
Classification & \texttt{page\_\allowbreak{}blocks} & +0.0001 & 0.6251 & +0.0012 & 0.9997 & Insensitive \\
Classification & \texttt{parkinsons} & +0.0005 & 0.7554 & +0.0004 & 0.7710 & Insensitive \\
Classification & \texttt{pediatric\_\allowbreak{}appendicitis} & -0.0013 & 0.0758 & -- & -- & Insensitive \\
Classification & \texttt{pen\_\allowbreak{}digits} & -0.0001 & 0.3328 & -0.0000 & 0.3256 & Insensitive \\
Classification & \texttt{phishing\_\allowbreak{}websites} & +0.0000 & 0.5562 & -0.0000 & 0.4411 & Insensitive \\
Classification & \texttt{poker\_\allowbreak{}hand} & -0.0027 & $< 1e^{-4}$ & -0.0034 & $< 1e^{-4}$ & Sensitive (both) \\
Classification & \texttt{polish\_\allowbreak{}companies\_\allowbreak{}bankruptcy} & +0.0000 & 0.6225 & -- & -- & Insensitive \\
Classification & \texttt{post\_\allowbreak{}operative\_\allowbreak{}patient} & +0.0001 & 0.5362 & +0.0000 & 0.5115 & Insensitive \\
Classification & \texttt{predictive\_\allowbreak{}maintenance} & -0.0024 & $< 1e^{-4}$ & -0.0016 & 0.0004 & Sensitive (both) \\
Classification & \texttt{primary\_\allowbreak{}tumor} & +0.0017 & 0.9950 & +0.0013 & 0.9707 & Insensitive \\
Classification & \texttt{raisin} & +0.0006 & 0.8606 & +0.0005 & 0.8196 & Insensitive \\
Classification & \texttt{raisin\_\allowbreak{}dataset} & -0.0009 & 0.0514 & -0.0012 & 0.0052 & Sensitive (corr-only) \\
Classification & \texttt{rice} & -0.0011 & 0.0368 & -0.0008 & 0.0726 & Insensitive \\
Classification & \texttt{room\_\allowbreak{}occupancy\_\allowbreak{}estimation} & -0.0022 & 0.0013 & -0.0030 & 0.0010 & Sensitive (both) \\
Classification & \texttt{servo} & -0.0004 & 0.0990 & -0.0007 & 0.0036 & Sensitive (both) \\
Classification & \texttt{shuttle\_\allowbreak{}landing\_\allowbreak{}control} & -0.0010 & 0.1451 & +0.0012 & 0.9140 & Insensitive \\
Classification & \texttt{skin\_\allowbreak{}segmentation} & -0.0019 & $< 1e^{-4}$ & -0.0021 & $< 1e^{-4}$ & Sensitive (both) \\
Classification & \texttt{solar\_\allowbreak{}flare} & -0.0007 & 0.1716 & -0.0015 & 0.0355 & Insensitive \\
Classification & \texttt{sonar} & +0.0007 & 0.9702 & -0.0000 & 0.4928 & Insensitive \\
Classification & \texttt{soybean\_\allowbreak{}large} & -0.0000 & 0.1196 & +0.0000 & 0.7017 & Insensitive \\
Classification & \texttt{soybean\_\allowbreak{}small} & +0.0004 & 0.6823 & +0.0024 & 0.9994 & Insensitive \\
Classification & \texttt{space\_\allowbreak{}shuttle\_\allowbreak{}oring} & +0.0000 & -- & +0.0000 & -- & Insensitive \\
Classification & \texttt{spambase} & +0.0029 & 1.0000 & +0.0011 & 0.9846 & Insensitive \\
Classification & \texttt{spect\_\allowbreak{}heart} & -0.0028 & 0.1240 & -0.0017 & 0.2489 & Insensitive \\
Classification & \texttt{spectf\_\allowbreak{}heart} & -0.0003 & 0.4461 & +0.0008 & 0.6299 & Insensitive \\
Classification & \texttt{statlog\_\allowbreak{}australian\_\allowbreak{}credit} & -0.0003 & 0.0393 & +0.0000 & 0.5028 & Insensitive \\
Classification & \texttt{statlog\_\allowbreak{}german\_\allowbreak{}credit} & +0.0000 & 0.6225 & +0.0000 & 0.6225 & Insensitive \\
Classification & \texttt{statlog\_\allowbreak{}heart} & -0.0021 & 0.0024 & -0.0008 & 0.1294 & Sensitive (uncorr-only) \\
Classification & \texttt{statlog\_\allowbreak{}image\_\allowbreak{}segmentation} & -0.0024 & 0.0008 & -0.0040 & $< 1e^{-4}$ & Sensitive (both) \\
Classification & \texttt{statlog\_\allowbreak{}landsat} & -0.0003 & 0.1784 & -0.0007 & 0.0112 & Sensitive (corr-only) \\
Classification & \texttt{statlog\_\allowbreak{}shuttle} & -0.0008 & 0.0230 & -0.0004 & 0.1356 & Sensitive (uncorr-only) \\
Classification & \texttt{statlog\_\allowbreak{}vehicle} & -0.0004 & 0.0842 & -0.0004 & 0.1355 & Insensitive \\
Classification & \texttt{steel\_\allowbreak{}plates\_\allowbreak{}faults} & +0.0000 & 0.5000 & +0.0000 & 0.5000 & Insensitive \\
Classification & \texttt{student\_\allowbreak{}academics\_\allowbreak{}performance} & +0.0004 & 0.7696 & +0.0001 & 0.5626 & Insensitive \\
Classification & \texttt{student\_\allowbreak{}performance} & -0.0000 & 0.4487 & +0.0002 & 0.9596 & Insensitive \\
Classification & \texttt{taiwanese\_\allowbreak{}bankruptcy} & +0.0000 & 0.7339 & +0.0000 & 0.5000 & Insensitive \\
Classification & \texttt{thoracic\_\allowbreak{}surgery} & +0.0000 & 0.5000 & +0.0001 & 0.8999 & Insensitive \\
Classification & \texttt{thyroid\_\allowbreak{}cancer\_\allowbreak{}recurrence} & +0.0000 & 0.5319 & +0.0004 & 0.7767 & Insensitive \\
Classification & \texttt{tic\_\allowbreak{}tac\_\allowbreak{}toe} & -0.0002 & 0.4207 & +0.0014 & 0.9489 & Insensitive \\
Classification & \texttt{user\_\allowbreak{}knowledge\_\allowbreak{}modeling} & -0.0037 & 0.0010 & -0.0038 & $< 1e^{-4}$ & Sensitive (both) \\
Classification & \texttt{vertebral\_\allowbreak{}column} & -0.0029 & 0.0079 & -0.0015 & 0.0986 & Insensitive \\
Classification & \texttt{voting\_\allowbreak{}records} & -0.0001 & 0.3331 & +0.0001 & 0.7549 & Insensitive \\
Classification & \texttt{waveform} & -0.0001 & 0.2656 & -0.0003 & 0.0547 & Insensitive \\
Classification & \texttt{website\_\allowbreak{}phishing} & -0.0005 & 0.2208 & -0.0012 & 0.0251 & Insensitive \\
Classification & \texttt{wine} & -0.0034 & 0.0300 & -0.0014 & 0.1969 & Insensitive \\
Classification & \texttt{wine\_\allowbreak{}quality} & -0.0002 & 0.2812 & -0.0003 & 0.1623 & Insensitive \\
Classification & \texttt{yeast} & -0.0003 & 0.1673 & -0.0008 & 0.0107 & Sensitive (corr-only) \\
Classification & \texttt{youtube\_\allowbreak{}spam} & -0.0006 & 0.3147 & -0.0020 & 0.0435 & Insensitive \\
Classification & \texttt{zoo} & -0.0033 & 0.0129 & +0.0016 & 0.9152 & Sensitive (uncorr-only) \\
Regression & \texttt{abalone} & -0.0021 & 0.0397 & +0.0016 & 0.8358 & Insensitive \\
Regression & \texttt{air\_\allowbreak{}quality} & +0.0113 & 0.9515 & -0.0027 & 0.1011 & Insensitive \\
Regression & \texttt{airfoil\_\allowbreak{}self\_\allowbreak{}noise} & +0.0005 & 0.7006 & +0.0002 & 0.7177 & Insensitive \\
Regression & \texttt{appliances\_\allowbreak{}energy} & -0.0003 & 0.1510 & -0.0050 & 0.1116 & Insensitive \\
Regression & \texttt{auto\_\allowbreak{}mpg} & -0.0062 & $< 1e^{-4}$ & -0.0015 & 0.2315 & Sensitive (uncorr-only) \\
Regression & \texttt{bike\_\allowbreak{}sharing} & -0.0005 & 0.2273 & +0.0014 & 0.9350 & Insensitive \\
Regression & \texttt{combined\_\allowbreak{}cycle\_\allowbreak{}power\_\allowbreak{}plant} & -0.0018 & 0.0934 & +0.0009 & 0.7330 & Insensitive \\
Regression & \texttt{computer\_\allowbreak{}hardware} & -0.0202 & 0.0018 & -0.0099 & 0.0893 & Sensitive (both) \\
Regression & \texttt{concrete\_\allowbreak{}compressive\_\allowbreak{}strength} & +0.0000 & 0.5064 & -0.0008 & 0.0383 & Insensitive \\
Regression & \texttt{concrete\_\allowbreak{}slump} & -0.0000 & 0.3550 & -0.0002 & 0.3203 & Insensitive \\
Regression & \texttt{energy\_\allowbreak{}efficiency} & -0.0023 & 0.0549 & -0.0028 & 0.0593 & Insensitive \\
Regression & \texttt{forest\_\allowbreak{}fires} & +0.0003 & 0.9083 & +0.0003 & 0.8783 & Insensitive \\
Regression & \texttt{gas\_\allowbreak{}turbine\_\allowbreak{}emission} & +0.0001 & 0.6587 & +0.0001 & 0.5411 & Insensitive \\
Regression & \texttt{heart\_\allowbreak{}failure} & -0.0004 & 0.2455 & -0.0002 & 0.1847 & Insensitive \\
Regression & \texttt{parkinson\_\allowbreak{}telemonitoring} & -0.0028 & 0.0002 & -0.0041 & 0.0071 & Sensitive (both) \\
Regression & \texttt{power\_\allowbreak{}consumption\_\allowbreak{}tetouan} & +0.0007 & 0.9513 & -0.0000 & 0.4605 & Insensitive \\
Regression & \texttt{seoul\_\allowbreak{}bike\_\allowbreak{}sharing} & -0.0033 & 0.0232 & -0.0017 & 0.1423 & Sensitive (uncorr-only) \\
Regression & \texttt{servo} & -0.0005 & 0.1800 & +0.0004 & 0.6813 & Insensitive \\
Regression & \texttt{student\_\allowbreak{}performance} & -0.0008 & 0.0918 & -0.0007 & 0.1364 & Insensitive \\
Regression & \texttt{superconductivity} & -0.0000 & 0.2025 & -0.0001 & 0.2004 & Sensitive (both) \\
Regression & \texttt{wine\_\allowbreak{}quality} & -0.0008 & 0.0941 & +0.0002 & 0.6447 & Insensitive \\
\end{longtable}
\endgroup

\subsection{Noise-induced datapoint shift visualization}
\begin{figure}[tbp]
\centering
\includegraphics[width=\linewidth]{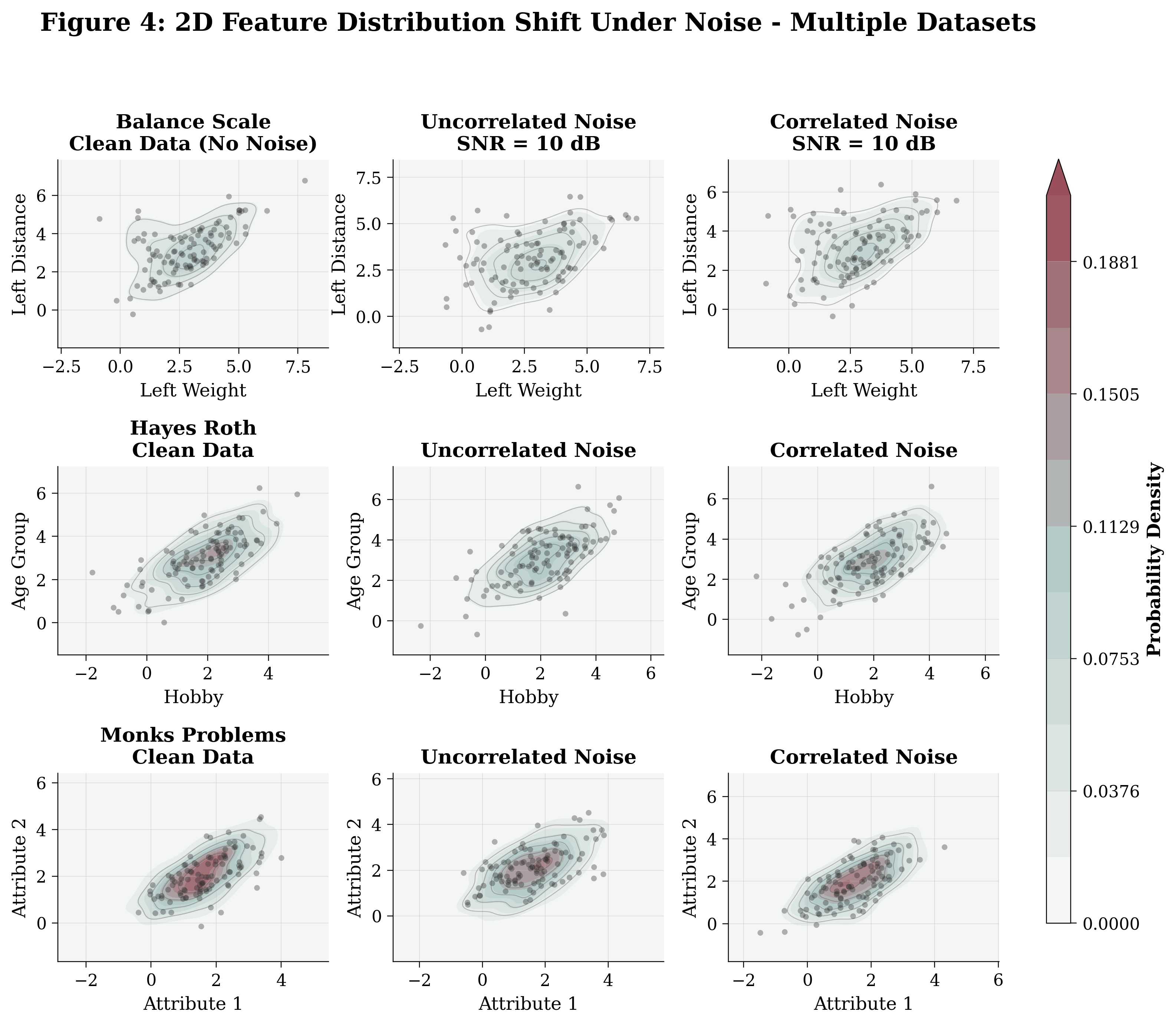}
\caption{2D datapoint and density-shift illustration under noise perturbations
Left: clean baseline. Middle: uncorrelated Gaussian noise.
Right: correlated Gaussian noise.}
\label{fig:appendix-noise-shift-2d}
\end{figure}

Figure~\ref{fig:appendix-noise-shift-2d} illustrates how perturbations shift point
clouds and density structure under increasing noise. Rows correspond to representative
datasets; columns compare clean data to uncorrelated and correlated noise at matched
SNR.

\subsection{Breakdown of UCI results by additional levels}
In \cref{tab:uci-sensitivity-analysis-breakdown} we present a breakdown of the results in our study broken down by several categories of interest. For the variety of datasets employed in our UCI study, there is little indication of domain or dataset size dependence. However, we do note possible indications of dependency relative to large sets of features, which may stem from our deliberate limitation of only employing 10 numerical features to mitigate challenges with token length. Similarly it is important to note prompt serialization, for which we have taken a simple approach of dataset description, metadata and covariate distributions in addition to raw numerical values, comparable to studies such as CLLM \citep{seedat2024curatedllmsynergyllms}.

\begin{table}[tbp]
\centering
\scriptsize
\renewcommand{\arraystretch}{1.2}
\setlength{\tabcolsep}{5pt}
\begin{tabular}{@{} lp{0.3cm} r r l rrr @{}}
\toprule
& & & & & \multicolumn{3}{c}{\textbf{Breakdown (Count)}} \\ \cmidrule(l){6-8}
\textbf{Factor / Level} & \textbf{Task} & \textbf{N} & \textbf{Sens.} & \textbf{Sensitive \%} & \textbf{Both} & \textbf{Corr.} & \textbf{Unc.} \\ 
\midrule

\rowcolor{gray!10} \multicolumn{8}{l}{\textit{Area of Application}} \\
Biology & C & 14 & 8 & 57.1 \rule{1.42cm}{1.5mm} & 3 & 3 & 2 \\
        & R & 1  & 0 & 0.0  & 0 & 0 & 0 \\
Business & C & 13 & 3 & 23.1 \rule{0.57cm}{1.5mm} & 2 & 1 & 0 \\
         & R & 2  & 1 & 50.0 \rule{1.25cm}{1.5mm} & 0 & 0 & 1 \\
Comp. Science & C & 20 & 6 & 30.0 \rule{0.75cm}{1.5mm} & 5 & 1 & 0 \\
               & R & 7  & 2 & 28.6 \rule{0.71cm}{1.5mm} & 2 & 0 & 0 \\
Health/Medicine & C & 43 & 17 & 39.5 \rule{0.98cm}{1.5mm} & 13 & 2 & 2 \\
                & R & 1  & 0 & 0.0  & 0 & 0 & 0 \\
Physics/Chem & C & 17 & 4 & 23.5 \rule{0.58cm}{1.5mm} & 2 & 1 & 1 \\
             & R & 3  & 1 & 33.3 \rule{0.83cm}{1.5mm} & 1 & 0 & 0 \\
Social Science & C & 14 & 5 & 35.7 \rule{0.89cm}{1.5mm} & 2 & 2 & 1 \\
               & R & 3  & 0 & 0.0  & 0 & 0 & 0 \\

\midrule
\rowcolor{gray!10} \multicolumn{8}{l}{\textit{Dataset Size (Samples)}} \\
$\le 500$    & C & 54 & 23 & 42.6 \rule{1.06cm}{1.5mm} & 17 & 3 & 3 \\
             & R & 4  & 2  & 50.0 \rule{1.25cm}{1.5mm} & 1 & 0 & 1 \\
$501-1,000$  & C & 20 & 8  & 40.0 \rule{1.00cm}{1.5mm} & 5 & 2 & 1 \\
$1,001-5,000$& C & 26 & 6  & 23.1 \rule{0.57cm}{1.5mm} & 4 & 2 & 0 \\
$> 5,000$    & C & 38 & 12 & 31.6 \rule{0.79cm}{1.5mm} & 6 & 4 & 2 \\
             & R & 10 & 3  & 30.0 \rule{0.75cm}{1.5mm} & 2 & 0 & 1 \\

\midrule
\rowcolor{gray!10} \multicolumn{8}{l}{\textit{Feature Count}} \\
$\le 10$     & C & 48 & 23 & 47.9 \rule{1.19cm}{1.5mm} & 18 & 4 & 1 \\
             & R & 9  & 2  & 22.2 \rule{0.55cm}{1.5mm} & 1 & 0 & 1 \\
$11-20$      & C & 40 & 18 & 45.0 \rule{1.12cm}{1.5mm} & 13 & 2 & 3 \\
$21-50$      & C & 33 & 7  & 21.2 \rule{0.53cm}{1.5mm} & 1 & 4 & 2 \\
$> 50$       & C & 17 & 1  & 5.9  \rule{0.14cm}{1.5mm} & 0 & 1 & 0 \\
             & R & 2  & 1  & 50.0 \rule{1.25cm}{1.5mm} & 1 & 0 & 0 \\

\midrule
\rowcolor{gray!10} \multicolumn{8}{l}{\textit{Class Count (Classification Only)}} \\
2       & C & 70 & 21 & 30.0 \rule{0.75cm}{1.5mm} & 12 & 6 & 3 \\
3--5    & C & 33 & 12 & 36.4 \rule{0.91cm}{1.5mm} & 10 & 1 & 1 \\
6--10   & C & 22 & 12 & 54.5 \rule{1.36cm}{1.5mm} & 7 & 3 & 2 \\
$> 10$  & C & 12 & 4  & 33.3 \rule{0.83cm}{1.5mm} & 3 & 1 & 0 \\

\bottomrule
\end{tabular}
\caption{Sensitivity analysis: Relationship between dataset characteristics and performance deterioration. Bars represent relative sensitivity (\%). (C: Classification, R: Regression)}
\label{tab:uci-sensitivity-analysis-breakdown}
\end{table}

\section{Implementation and Reproducibility Details}
\subsection{IMDB lexical perturbation progression}
\begin{figure}[t]
\centering
\scriptsize
\begin{tikzpicture}[node distance=6mm] %
\tikzset{
  stagebase/.style={draw=gray!65, line width=0.6pt, rounded corners=2mm, text width=0.92\linewidth, align=left, inner sep=5pt},
  stageclean/.style={stagebase, fill=gray!5},
  stagelight/.style={stagebase, fill=gray!9},
  stagemed/.style={stagebase, fill=gray!13},
  stagemax/.style={stagebase, fill=gray!17}
}

\node[stageclean] (s0) {\textbf{Severity 0.00 (clean)}\\{\ttfamily I rank OPERA as one of the better Argento films. Plot holes and inconsistencies? Sure, but I don't think they impair this film as much as many other reviewers seem to. A lot of elements that are in many of}};

\node[stagelight, below=of s0] (s1) {\textbf{Severity 0.33 (light lexical noise)}\\{\ttfamily Ir ank [MASK] npe npe [MASK] bette [MASK] iflms. [MASK] oel snad inconsistenicesb? utI tihnk imapirt his fiml a ass many [MASK] reviewers [MASK] A elemenst th atare in of}};

\node[stagemed, below=of s1] (s2) {\textbf{Severity 0.67 (medium lexical noise)}\\{\ttfamily I rank [MASK] yoles [MASK] iconsitsneices? Sure ,It hnik hnik film film [MASK] as as A [MASK]}};

\node[stagemax, below=of s2] (s3) [below=of s2] {\textbf{Severity 1.00 (maximum lexical noise)}\\{\ttfamily [MASK] hte and [MASK] [MASK] [MASK] kto [MASK]}};

\draw[->, gray!60, line width=0.8pt] (s0.south) -- (s1.north);
\draw[->, gray!60, line width=0.8pt] (s1.south) -- (s2.north);
\draw[->, gray!60, line width=0.8pt] (s2.south) -- (s3.north);
\end{tikzpicture}
\caption{Lexical perturbation progression for a random IMDB sample
(\texttt{test\_022423}, label: positive), generated with the same corruption pipeline
used in the experiments. As severity increases, token-level corruption and masking rapidly
erase lexical signal.}
\label{fig:appendix-imdb-lexical-progression}
\end{figure}
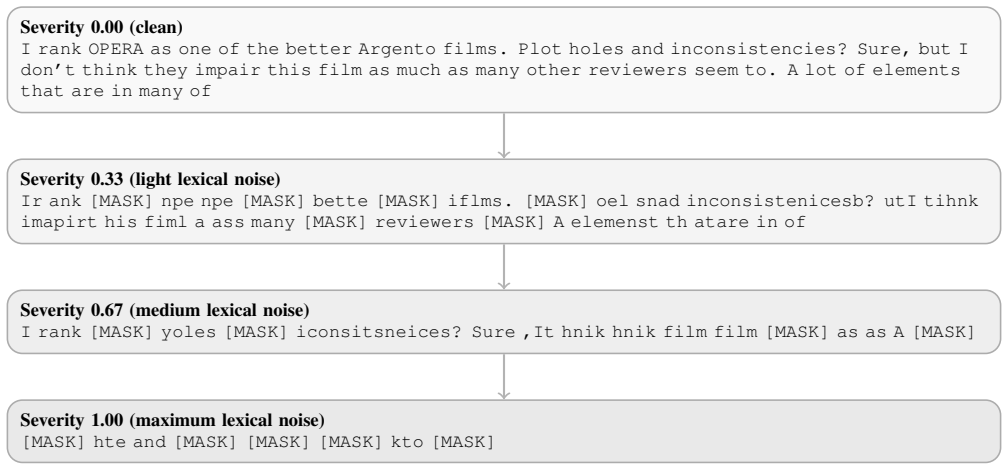

In \cref{fig:appendix-imdb-lexical-progression} below we show a single IMDB review example
and the resulting lexical corruption sequence at four severities from 0 to 1.

\clearpage
\subsection{Prompt construction and run setup}\label{box:prompt-setup}
\begin{tcolorbox}[
  enhanced,
  breakable,
  colback=gray!3,
  colframe=gray!55,
  boxrule=0.6pt,
  arc=2mm,
  left=6pt,
  right=6pt,
  top=6pt,
  bottom=6pt,
  title={Prompt template \& Execution},
  fonttitle=\bfseries
]
\textbf{System prompt} \\
Each request starts with dynamic dataset context, task definition, label space (or regression
target), and explicit output-format constraints. \\

\textbf{User prompt} \\
The user message contains few-shot examples as $(X, y)$ pairs, followed by evaluation
rows as $X$ only. For tabular settings, inputs are restricted to at most 10 numeric
features to control context length. \\

\textbf{Run protocol} \\
Each dataset $\times$ noise setting is run for five independent trials using fixed random
seeds. Text experiments apply the lexical corruption at
the requested severity; tabular experiments apply Gaussian perturbations under correlated
or uncorrelated covariance structure. \\

\textbf{Output schema} \\
Model outputs are schema-validated against task-specific constraints (label set for
classification, numeric parsing for regression). Invalid schema/parse outputs are retried up to 3 times; otherwise counted missing.
\end{tcolorbox}

\section{Acknowledgements}
We thank Jan Ole Ernst and Lars Holdijk for helpful discussions and thoughtful
conversations related to this work.

\end{document}